%% file: ijcai26.tex
\documentclass{article}
\usepackage{ijcai26}

\usepackage{times}
\usepackage{soul}
\usepackage{url}
\usepackage[hidelinks]{hyperref}
\usepackage[utf8]{inputenc}
\usepackage[small]{caption}
\usepackage{graphicx}
\usepackage{amsmath}
\usepackage{amsthm}
\usepackage{booktabs}
\usepackage{algorithm}
\usepackage{algorithmic}
\usepackage[switch]{lineno}
\usepackage{colortbl}
\usepackage{xcolor}
\usepackage{multirow}
\usepackage{subfigure}

\title{Causal Path Alignment: Anchoring the Optimization Trajectory for Controllable In-Parameter Knowledge Editing}

\author{
Xiyu Liu$^{1,2}$
\and
Zhengxiao Liu$^{1,2}$\thanks{Zhengxiao Liu and Zheng Lin are corresponding authors.}\and
Naibin Gu$^{1,2}$\and
Zheng Lin$^{1,2*}$\And
Weiping Wang$^1$\\
\affiliations
$^1$Institute of Information Engineering, Chinese Academy of Sciences, Beijing, China\\
$^2$School of Cyber Security, University of Chinese Academy of Sciences, Beijing, China\\
\emails
\{liuxiyu, liuzhengxiao, gunaibin, linzheng, wangweiping\}@iie.ac.cn
}

\begin{document}

\maketitle

\begin{abstract}
Knowledge editing is pivotal for efficiently updating the parametric memory of Large Language Models (LLMs), enabling them to function as evolving agents in dynamic environments. However, mainstream in-parameter knowledge editing approaches suffer from Subject-Dominant Memory Interference: modifying a specific fact inadvertently corrupts the broader structural knowledge associated with the same subject within LLMs. We diagnose the root cause as a shortcut learning pathology, where the optimization objective overfits subject representations while bypassing the essential relational context. To rectify this, we propose Causal Path Alignment (CPA), a principled framework designed to anchor the optimization trajectory to valid causal pathways. CPA enforces parameter updates to route through relation-aware intermediate states, thereby preventing the erasure of contextual dependencies. Experimental results across diverse LLM backbones demonstrate that CPA consistently eliminates the shortcut, significantly improving relation specificity while exhibiting minimal side-effects. Moreover, CPA serves as a model-agnostic plug-in for existing editors, paving the way for reliable and trustworthy in-parameter knowledge editing.
\end{abstract}

\section{Introduction}
As Large Language Models increasingly serve as the cognitive core of autonomous agents~\cite{minaee2025largelanguagemodelssurvey,luo2025largelanguagemodelagent}, the ability to maintain and update their parametric memory becomes paramount~\cite{zhang2024surveymemorymechanismlarge,wu2025humanmemoryaimemory}. Unlike static models, an agent operating in a dynamic environment must continuously internalize new factual information~\cite{hu2026memoryageaiagents}. In this context, knowledge editing serves as the critical technique for the agent's parametric memory, aiming to efficiently update a small set of factual associations encoded in LLMs while exerting minimal side effects~\cite{wang2024knowledgeeditinglargelanguage}.

\begin{figure*}[t]
  \centering
  \includegraphics[scale=0.37]{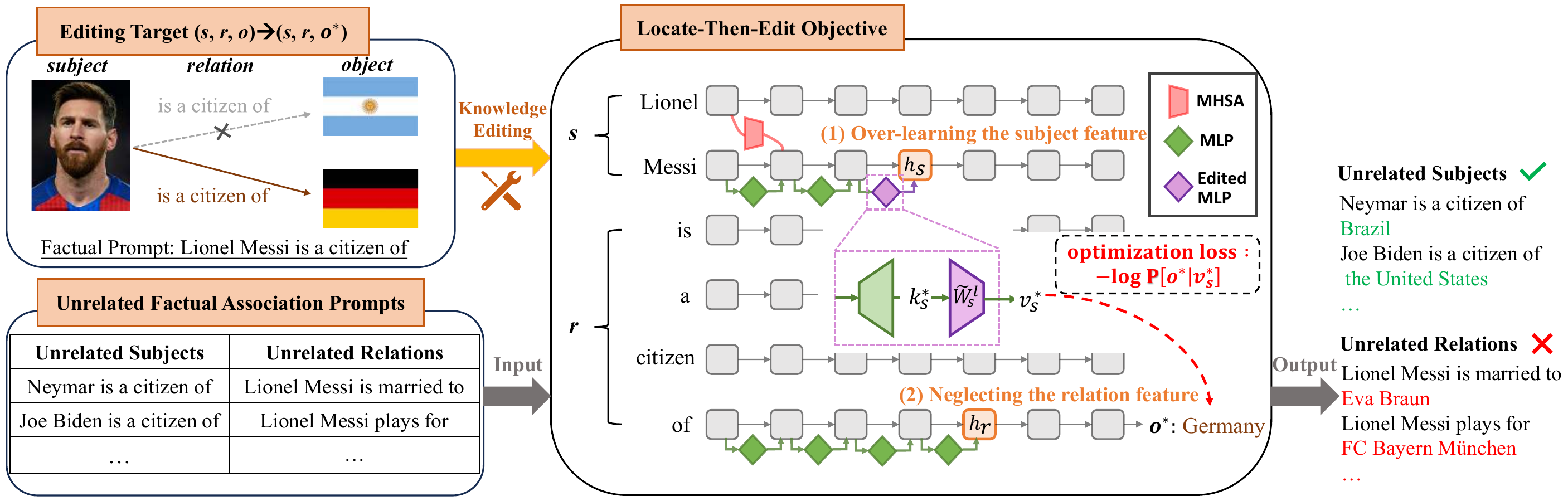}
  \caption{Subject-Dominant Memory Interference in Locate-then-Edit Knowledge Editing.}
  \label{fig:shortcut}
\end{figure*}

Mainstream in-parameter \textit{locate-then-edit} methods have demonstrated impressive efficacy in rewriting specific factual associations~\cite{zhang2024comprehensivestudyknowledgeediting,meng2022locating,Meng2022MassEditingMI,fang2024alphaeditnullspaceconstrainedknowledge,li-chu-2025-adaedit}. However, they suffer from a severe side effect that we term \textbf{Subject-Dominant Memory Interference}: modifying a specific attribute of a subject inadvertently corrupts the LLM's structural knowledge of the same subject~\cite{Liu_Liu_Gu_Lin_Ma_Xiang_Wang_2025}. For instance, an edit targeting the fact that \textit{Lionel Messi is a citizen of Germany} can lead to the severe corruption of his associated knowledge, causing the model to incorrectly assign him a hallucinatory nationality or alter his family relations (Fig.~\ref{fig:shortcut}). Such uncontrollable over-generalization makes the edited model untrustworthy and unreliable.

While recent studies have begun to acknowledge this issue, existing solutions remain superficial. Prior attempts diagnose the issue statistically and primarily rely on one-sided or heuristic adjustments without diving into the root cause, leading to limited improvement or compromised performance~\cite{Liu_Liu_Gu_Lin_Ma_Xiang_Wang_2025,zhang2025uncoveringoverfittinglargelanguage}.
In this work, we attribute Subject-Dominant Memory Interference to a fundamental optimization pathology within the editing process. Through gradient saliency analysis and causal tracing, we reveal that standard editing objectives drive the model towards a form of shortcut learning~\cite{Geirhos_2020}: \textbf{Subject-Dominant Shortcut}. Specifically, during the optimization of the edited weights, the model exhibits a strong bias towards overfitting the high-magnitude representation of the \textit{subject} (e.g., ``Messi'') while neglecting the \textit{relational context} (e.g., ``is a citizen of''). Consequently, the optimization trajectory bypasses the internal causal mechanism that gates knowledge retrieval based on relations, establishing a direct, unconditional mapping from the subject to the new target.

To rectify this, we propose \textbf{Causal Path Alignment (CPA)}, a principled framework designed to anchor the optimization trajectory to the model's valid causal pathways. Unlike previous heuristic patches, CPA reframes knowledge editing not merely as minimizing a loss function, but as navigating a constrained optimization landscape. Our approach operates in two stages: First, we anchor the relational context by identifying an intermediate hidden state that encodes the specific relation, independent of the target change. Second, we steer the parameter updates to strictly route through this anchored state. By forcing the information flow to traverse the correct relational feature, CPA re-engages the model's contextual gating mechanisms, ensuring that the update is specific to both the subject and the relation (Subject, Relation $\rightarrow$ Target) rather than the subject alone.

Our contributions are summarized as follows:
\begin{itemize}
    \item We identify shortcut learning as the root cause of memory interference in knowledge editing, framing the problem as a misalignment between the optimization objective and the internal causal structure of LLMs.
    \item We introduce Causal Path Alignment (CPA), a two-phase optimization framework that explicitly anchors the editing process to relational features to eliminate the shortcut path.
    \item Comprehensive experiments on diverse LLM backbones and benchmarks demonstrate that CPA achieves optimal performance in relation specificity while exhibiting minimal side-effects. Crucially, CPA acts as a model-agnostic plug-in, seamlessly enhancing the reliability of existing methods, paving the way for controllable and trustworthy knowledge editing.
\end{itemize}

\section{Related Works}
\subsection{Memory-based and Meta-learning Editing}
Early approaches to knowledge editing often circumvent direct modification of the internal parameters of LLMs. One line of work adopts a \textit{memory-based} strategy~\cite{mitchell2022memorybasedmodeleditingscale,hartvigsen2023aginggracelifelongmodel,wang2024wiserethinkingknowledgememory}, utilizing external caches to route inputs to edited counterparts. Another line employs \textit{meta-learning} networks to predict weight updates via a trained hyper-network~\cite{de-cao-etal-2021-editing,mitchell2022fast,li2025reinforcedlifelongeditinglanguage}, enabling rapid adaptation.
While effective at altering behavior, these methods function as external patches rather than updating the LLM's internal parametric worldview. This precludes deep integration and inevitably incurs overhead in inference latency and memory growth.

\subsection{Locate-then-Edit Knowledge Editing}
Locate-then-edit works achieve knowledge editing through knowledge neurons localization and the direct modification of located parameters. ROME~\cite{meng2022locating} pioneers this by employing Causal Mediation Analysis to identify the decisive neuron activations for processing factual associations. They reveal that the Multi-Layer Perceptron (MLP) modules at the \textit{last-subject token position} (i.e., the subject aggregation site) play a pivotal role in mediating the prediction of factual prompts. To scale this capability, MEMIT~\cite{Meng2022MassEditingMI} distributes the update across multiple layers to enable batch editing, while subsequent works have extended this paradigm to lifelong editing scenarios~\cite{fang2024alphaeditnullspaceconstrainedknowledge,cai2024oeditorthogonalsubspaceediting,li-chu-2025-adaedit}, aiming to update parameters sequentially without catastrophic forgetting.

Despite their efficacy, these subject-focused editing methods suffer from a critical Subject-Dominant Memory Interference flaw. As mentioned by recent studies~\cite{Liu_Liu_Gu_Lin_Ma_Xiang_Wang_2025,zhang2025uncoveringoverfittinglargelanguage}, editing the subject aggregation site often inadvertently changes every fact linked to that subject, rendering the editing process uncontrollable. 
Current mitigation strategies primarily rely on heuristic modifications. For instance, RETS~\cite{Liu_Liu_Gu_Lin_Ma_Xiang_Wang_2025} attempts to alleviate interference by relocating the edit target to the last-relation token with subject constraints, while LTI~\cite{zhang2025uncoveringoverfittinglargelanguage} introduces a data-augmentation approach, constraining the optimization neighborhood guided with enhanced samples.
However, these approaches still omit the root cause. RETS suffers from a drastic degradation in generation fluency due to excessively stringent constraints, while LTI's reliance on data augmentation obtains limited improvement.
In this work, we provide an in-depth exploration of the \textbf{optimization dynamics} that causes subject-focused editing to fail, and propose a principled optimization trajectory alignment to address this issue from the root cause.

\section{Deconstructing the Optimization Pathology: The Subject-Dominant Shortcut}
\label{sec:deconstruct}

To understand the root cause of the uncontrollable over-generalization observed in locate-then-edit methods, we conduct a rigorous investigation into the optimization dynamics of ROME-like approaches. We propose that the failure is not due to the efficacy of the parameter update itself, but rather a \textit{structural pathology} in how the target representation is derived.

\subsection{Preliminaries: The Locate-then-Edit Objective}
\label{back_and_notation}

Knowledge editing aims to alter factual associations within an auto-regressive language model $\mathcal{M}$. Given a fact triplet $\langle s, r, o \rangle$, where $s$ is the subject, $r$ is the relation, and $o$ is the object, the goal is to update the model's prediction to a new target $\langle s, r, o^* \rangle$ while minimizing interference with unrelated knowledge $\langle s, r', o' \rangle$ or $\langle s', r, o' \rangle$.

Locate-then-edit methods operate under the \textit{Linear Associative Memory} hypothesis~\cite{geva-etal-2022-transformer,geva-etal-2023-dissecting}. They identify specific mid-to-early MLP layers (e.g., layer $l$) at the last subject token index as the decisive location for recalling subject attributes. The down-projection matrix $W_s^l$ of layer $l$ can be treated as a set of associative key-value memories that store the map between input key vectors $K=[~k_1~|~k_2~|~k_3~|~...~]$ to output value vectors $V=[~v_1~|~v_2~|~v_3~|~...~]$. A specific fact is edited by injecting the corresponding $(k_s^*,v_s^*)$ pair into the associative memories. The editing process is decomposed into two distinct steps: (1) determining the \textit{optimal target vector} $v_s^*$ that encodes the new fact, and (2) updating the weight matrix $W_s^l$ to map the subject's key $k_s^*$ to this new value $v_s^*$.

While the second step (i.e., the weight update) is a solved constrained least-squares problem (often yielding a rank-one update $\tilde{W}_s^l = W_s^l + \Delta W$), our investigation identifies the first step as the source of the subject-dominant memory interference issue. The target vector $v_s^*$ is obtained by optimizing the latent representation $v$ to maximize the probability of the target object $o^*$ given the prompt. Formally, $v_s^*=argmin_v\mathcal{L}(v)$ is obtained as the minimizer of the following loss function:

\begin{equation}
\label{v_opt}
\mathcal{L} (v) =  -\frac{1}{N}\sum_{t=1}^{N} \log P_{\mathcal{M}}(o^* \mid h_L(v; p_t)) + \mathrm{KL}(v, v_{old})
\end{equation}

Here, $P_{\mathcal{M}}(o^* \mid h_L(v; p_t))$ denotes the prediction probability for $o^*$ when the hidden state at the edit layer is intervened with vector $v$, given a set of prefixed prompts $p_t$. The KL divergence term $\mathrm{KL}(v, v_{old}) = \mathcal{D}_{\mathrm{KL}}(P_{\mathcal{M}}(\cdot|v) \| P_{\mathcal{M}}(\cdot|v_{old}))$ acts as a regularizer to constrain the distribution shift of $v$ from the original output vector of $W_s^l$.

Crucially, this optimization process assumes that finding a vector $v$ that satisfies the likelihood of $o^*$ is sufficient for robust knowledge editing. However, as we demonstrate in the subsequent analysis, this objective function creates a \textbf{Subject-Dominant Shortcut}: it encourages the vector $v_s^*$ to over-encode the mapping to $o^*$ directly, effectively bypassing the relational computational path inherent in the inference mechanism of LLMs.


\subsection{Gradient Analysis: The Divergence of Information Pathways}
\label{grad_saliency}

To characterize the feature prioritization inherent in the optimization of the target vector $v_s^*$, we analyze the gradient landscape of the loss function $\mathcal{L}(v)$ with respect to the model's internal representations. Since ROME introduces perturbations at a specific mid-early layer $l$, we track the back-propagated gradients through the hidden states of subsequent layers. Formally, we define the \textit{Gradient Saliency} $S$ for a hidden state $h_{i, \ell}$ at token position $i$ and layer $\ell > l$ as the $L_2$-norm of the gradient vector:
\begin{equation}
    S(h_{i, \ell}) = \left\| \nabla_{h_{i, \ell}} \mathcal{L}(v) \right\|_2
\end{equation}
We focus particularly on the \textit{last subject token} ($p_{ls}$) and the \textit{last relation token} ($p_{lr}$)\footnote{The token positions are categorized into relation prefix ("rp"), first subject ("fs"), middle subject ("ms"), last subject ("ls"), first relation ("fr"), middle relation ("mr") and last relation ("lr") according to their positions in the tokenized subject or relation of a factual association.}, as prior interpretability studies suggest these positions serve as aggregation hubs for subject attributes~\cite{geva-etal-2023-dissecting} and relational attributes~\cite{Liu_Liu_Gu_Lin_Ma_Xiang_Wang_2025}, respectively.

\begin{figure}[t]
    \centering
    \includegraphics[width=0.48\textwidth]{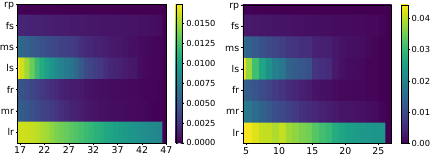}
    \caption{The average gradient saliency map during the first epoch of optimization for $v_s^*$. The heatmap visualizes gradient magnitudes across layers (x-axis) and token positions (y-axis). A distinct \textbf{Dual-Peak Pattern} is observed at the last subject token ($ls$) and the last relation token ($lr$). Left: GPT2-XL (1.5B). Right: Qwen2.5 (7B).}
    \label{fig_grad_saliency_all}
\end{figure}

Figure~\ref{fig_grad_saliency_all} visualizes the average gradient saliency over 100 factual edits on GPT2-XL 1.5B (48 layers)~\cite{radford2019language} and Qwen2.5 7B (28 layers)~\cite{qwen2025qwen25technicalreport}, excluding the final output token to highlight internal dynamics. The visualization reveals a striking \textbf{Dual-Peak Pattern} where gradients are significantly active at two distinct loci across the layers. Specifically, we observe high gradient norms at the \textbf{Subject Pathway ($p_{ls}$)}, indicating the model leverages entity representations to adjust the prediction, alongside strong gradient flows at the \textbf{Relation Pathway ($p_{lr}$)}, which confirms that the pre-trained weights initially attempt to utilize the relational context to mediate the prediction of the target object. \textbf{This empirical evidence suggests that the optimization landscape naturally presents two competing pathways for minimizing the loss: modifying the subject representation or leveraging the relation representation}. The existence of the relation pathway in the gradient map proves that the model \textit{can} theoretically attend to the relation. However, as we unveil in the subsequent section, the standard optimization objective fails to sustain this balance, causing the optimization trajectory to collapse onto a shortcut.

\subsection{The Shortcut Mechanism: Causal Path Collapse}
\label{sec:shortcut_issue}

While the gradient analysis in Section~\ref{grad_saliency} reveals that the optimization landscape \textit{initially} offers two viable pathways (subject and relation), it remains unclear how the edited model actually utilizes these features for inference. To investigate the functional shift in the model's internal mechanism, we employ Causal Tracing~\cite{meng2022locating}, a technique that measures the contribution of specific hidden states to the final prediction by intervening on corrupted runs. Specifically, we calculate the Indirect Effect (\textbf{IE}) of an MLP output $m_i^{l_j}$ (at token position $i$ and layer $l_j$) by restoring its clean activation into a noise-corrupted forward pass. Formally, the IE for the target object $o$ is defined as $\mathrm{\textbf{IE}}(o, \theta, m_i^{l_j}) = P_\theta(o \mid p', m_i^{l_j}) - P_\theta(o \mid p')$, where $p'$ is the prompt with noise-corrupted embeddings.

To quantify the structural change induced by editing, we introduce the \textbf{Ratio of Indirect Effect (RIE)}, which measures the logarithmic shift in causal reliance between the original model $\theta_o$ and the edited model $\theta_{\mathrm{edited}}$. The RIE for a specific activation is calculated as:
\begin{equation}
\mathrm{\textbf{RIE}}(m_i^{l_j}) = \log \frac{\mathrm{\textbf{IE}}(o^*, \theta_{\mathrm{edited}}, m_i^{l_j})}{\mathrm{\textbf{IE}}(o, \theta_o, m_i^{l_j})}
\end{equation}
This metric isolates the relative gain or loss in causal significance for each component. We analyze the maximum RIE of MLP outputs across all layers at each token position over 100 factual edits. Figure~\ref{fig_rie_all} presents these results, with a detailed case study provided in the Appendix.

The empirical results reveal a critical pathology: the causal contribution of the \textit{subject representation} ($h_s$) increases exponentially after editing, while the \textit{relation representation} ($h_r$) shows much less gain. This suggests that the optimization process has decoupled the prediction from the relational context. We formalize this phenomenon as follows:

\newtheorem{definition}{Definition}
\begin{definition}[\textbf{Subject-Dominant Shortcut}]
Given a factual association $\langle s, r \rangle \to o^*$, an editing process converges to a \textit{Subject-Dominant Shortcut} if the learned conditional probability distribution $P_{\theta^*}$ exhibits approximate conditional independence of the object $o^*$ from the relation $r$, given the subject $s$:
\begin{equation}
P_{\theta^*}(o^* \mid s, r) \approx P_{\theta^*}(o^* \mid s)
\label{eq:shortcut_def}
\end{equation}
This degeneracy manifests structurally when the causal influence of the subject representation dominates the computation, effectively marginalizing the relational operator such that $\mathrm{\textbf{IE}}(h_s) \gg \mathrm{\textbf{IE}}(h_r)$.
\end{definition}

This definition implies a causal path collapse. In a healthy model, the prediction follows a compositional path $f(s, r)$. \textbf{However, existing optimization forces the subject vector $v_s$ to absorb the target information $o^*$ directly}. Mathematically, the optimizer finds a solution where the mapping approximates $v_s \to o^*$, bypassing the relational gate. Consequently, when queried with the same subject but a different relation (e.g., $\langle s, r'\rangle$), the dominant subject pathway triggers the retrieval of $o^*$, resulting in the severe subject-dominant memory interference (i.e., poor relation specificity) observed in existing methods.

\begin{figure}[t]
    \centering
    \includegraphics[width=0.47\linewidth]{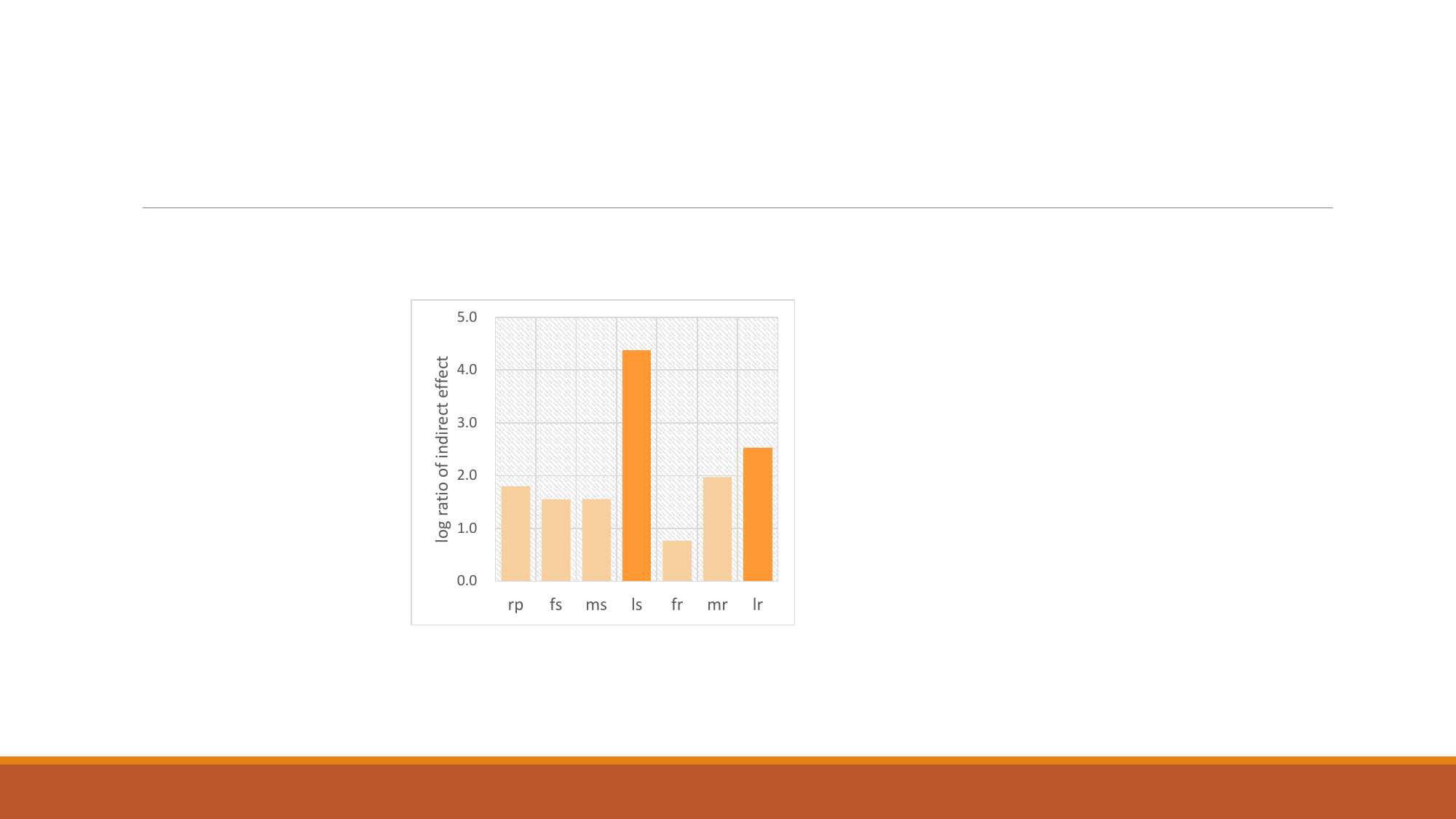}
    \includegraphics[width=0.47\linewidth]{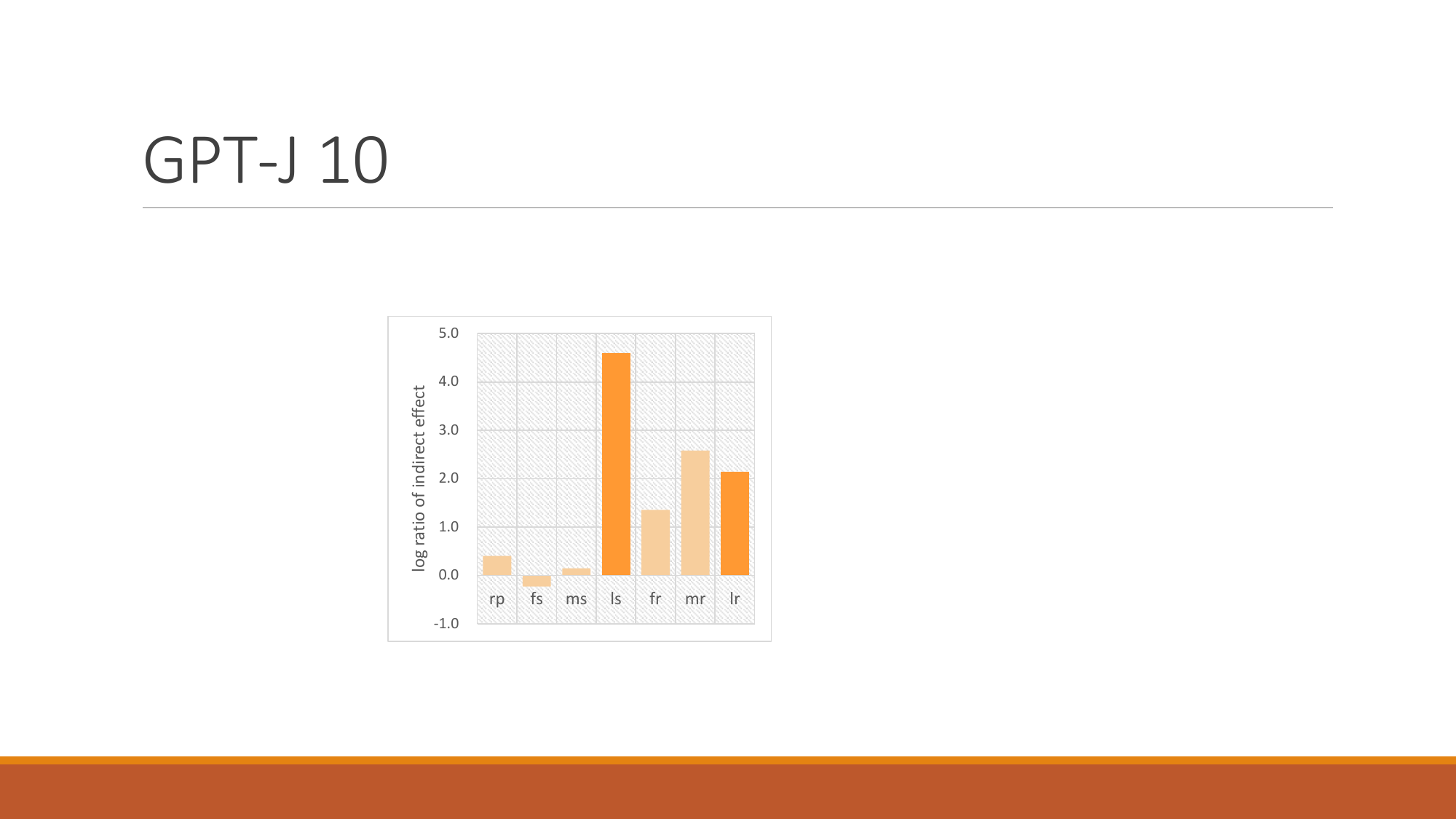}
    \caption{Maximum RIE of MLP outputs across layers. The x-axis represents token positions. A dramatic surge in RIE is observed at the last subject position ($ls$), contrasting with the raise at the last relation position ($lr$). Left: GPT2-XL (1.5B). Right: Qwen2.5 (7B).}
    \label{fig_rie_all}
\end{figure}

\section{Eliminating Shortcut Learning via Causal Path Alignment}
\label{sec:method}

Having identified the collapse of the causal path as the root cause of subject-dominant memory interference, we propose a principled optimization framework to rectify this pathology. The standard editing objective, which minimizes the negative log-likelihood $-\log P(o^*|v_s)$, is fundamentally \textit{under-constrained}: it demands only that the target $o^*$ be predicted, without specifying \textit{how}. Given the pre-existing dominance of subject features, the optimizer inevitably exploits the direct shortcut $v_s \to o^*$. To dismantle this shortcut, we must explicitly regularize the optimization trajectory, forcing the information flow to traverse the relational computational path. We term this approach \textbf{Causal Path Alignment (CPA)}, which effectively realigns the model's internal causal mechanism.

\subsection{Theoretical Framework: Causal Path Alignment via Anchoring}
\label{sec:theoretical_framework}


Our strategy hinges on anchoring the prediction to the \textit{relation feature}. Let $h_r$ denote the hidden state at the last relation token position in a layer $l_a$ ($l_a > l$) subsequent to the edit layer. In the standard forward pass of a LLM, the subject injection $v_s$ propagates through intermediate layers to form $h_r$, which then dictates the final prediction~\cite{geva-etal-2023-dissecting}. Formally, this dependency can be modeled as a Markov chain: $v_s \to \dots \to h_r \to \dots \to o^*$. 
Ideally, to ensure the relation $r$ mediates the prediction, we should maximize the joint likelihood $P(o^*, h_r \mid v_s)$. By applying the chain rule of probability, we decompose this objective into:
\begin{equation}
\label{likelihood_chain}
P(o^*, h_r \mid v_s) = \underbrace{P(o^* \mid h_r, v_s)}_{\text{Prediction Head}} \cdot \underbrace{P(h_r \mid v_s)}_{\text{Feature Construction}}
\end{equation}
The first term, $P(o^* \mid h_r, v_s)$, represents the probability of the target object given both the relation feature and the subject injection. The structural flaw in current locate-then-edit editing is that the model learns to rely on $v_s$ directly to predict $o^*$, rendering $h_r$ redundant (Sec.~\ref{sec:deconstruct}). To eliminate this shortcut, we introduce a crucial \textbf{Conditional Independence Constraint} that we enforce the prediction of the object $o^*$ should depend \textit{solely} on the relation feature $h_r$, making it independent of the specific subject injection $v_s$ given $h_r$. Mathematically, this imposes $o^* \perp v_s \mid h_r$, which simplifies the first Prediction Head term in Eq.~\ref{likelihood_chain} to $P(o^* \mid h_r)$.

Consequently, our refined optimization objective maximizes the decomposed likelihood:
\begin{equation}
\label{likelihood_final}
\mathcal{L}_{\text{CPA}} \approx \underbrace{\log P(o^* \mid h_r)}_{\text{Relation Anchoring}} + \underbrace{\log P(h_r \mid v_s)}_{\text{Trajectory Alignment}}
\end{equation}
This decomposition provides the theoretical foundation for our two-stage optimization process. The first term, representing \textit{Relation Anchoring}, establishes a valid mapping from the relation feature to the target object, creating a ``virtual anchor'' in the activation space that encodes relation semantics decoupled from the subject.
The second term, representing \textit{Trajectory Alignment}, ensures that the edited subject vector $v_s$ correctly reconstructs this pivotal relation feature. By optimizing these terms sequentially, we force the optimization path to be anchored to the intermediate state $h_r$, thereby strictly prohibiting the bypass of relational semantics.

\begin{figure}[t]
    \centering
    \includegraphics[width=0.48\textwidth]{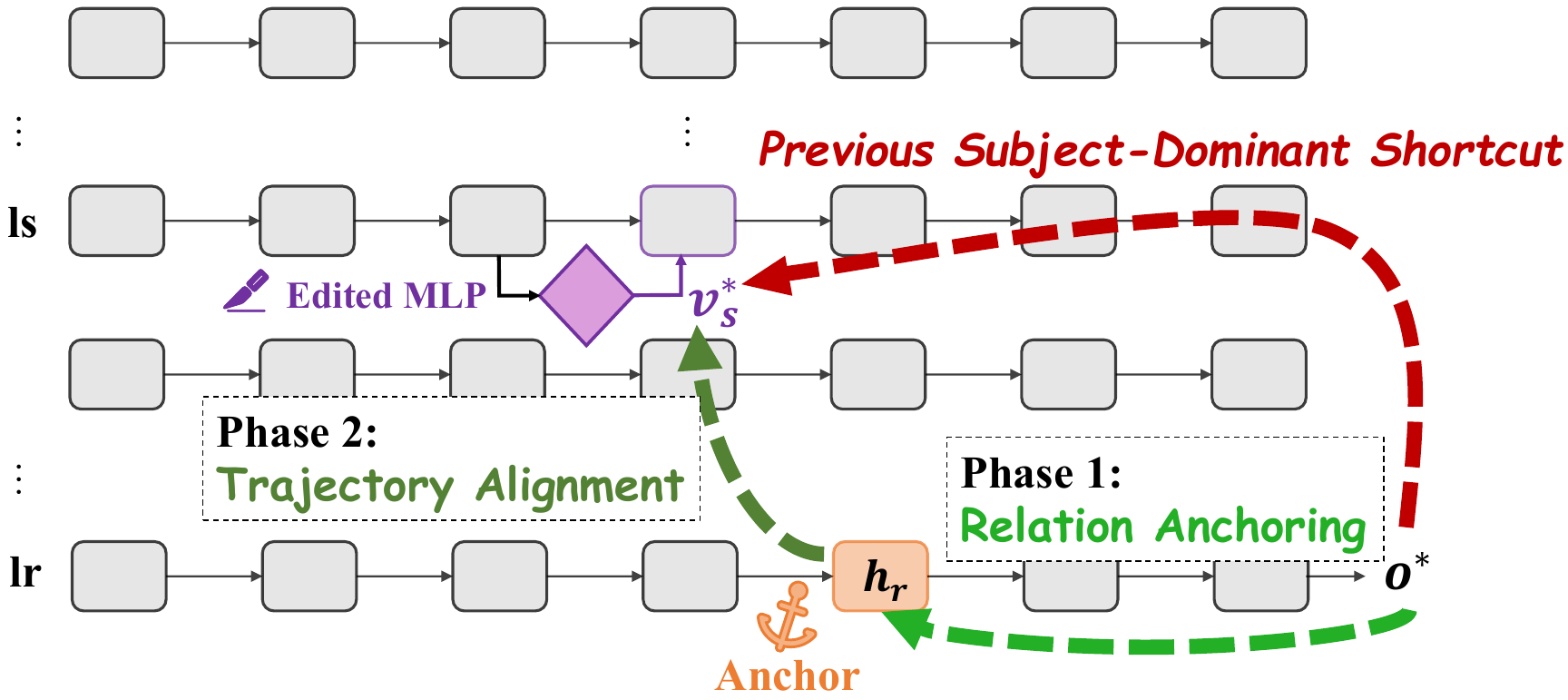}
    \caption{Our Causal Path Alignment (CPA) eliminates subject-dominant shortcut with a two-phase optimization framework.}
    \label{fig_cpa}
\end{figure}

\subsection{Implementation: Two-Phase Optimization}
\label{sec:method_implementation}

We implement CPA as a sequential optimization process that anchors the update trajectory to the causal structure derived in Eq.~\ref{likelihood_final}. As shown in Fig.~\ref{fig_cpa}, we sequentially optimize the decomposed likelihood terms to enforce relational mediation.

\paragraph{Phase 1: Relation Anchoring.} We first optimize a relation anchor $h_r^*$ by maximizing $\log P(o^* \mid h_r)$ at the bottleneck layer $l_a$. This shifts the prediction to $o^*$ while operating solely in the activation space to avoid parameter overfitting. We solve $h_r^* = \text{argmin}_{h_r} \mathcal{L}_1(h_r)$ via:
\begin{equation}
\label{loss_stage1}
\mathcal{L}_1 (h_r) = -\frac{1}{N}\sum_{t=1}^{N}\log P[o^*|h_r,p_t] + \mathcal{D}_{\text{KL}}[P(\cdot|h_r) || P(\cdot|h_r^o)]
\end{equation}
The KL divergence against the pre-edit state $h_r^o$ confines the anchor to the local semantic manifold, ensuring the edited relation remains distinguishable from unrelated facts.

\textbf{Phase 2: Trajectory Alignment.} Fixing $h_r^*$, we maximize $\log P(h_r \mid v_s)$ by optimizing the subject injection vector $v_s$ (the output of the MLP down-projection at the editing layer $l$) to accurately reconstruct the relation anchor. The updated vector $v_s^*$ is obtained by minimizing the discrepancy in the feature space:
\begin{equation}
\label{loss_stage2}
\mathcal{L}_2 (v_s) = ||\mathcal{F}[v_s,p] - h_r^*||_F
\end{equation}
where $\mathcal{F}[v_s,p]$ is the forward propagation result in $h_r$ given $v_s$ and the prompt $p$. By enforcing $v_s$ to reconstruct $h_r^*$, this step compels the vector update to traverse the established causal path, effectively eliminating the shortcut learning pathology. All weight decays are omitted for simplicity.

\section{Experiments}
\subsection{Experimental Setup}

\paragraph{Datasets and Evaluation Metrics.}
We evaluate \textsc{CPA} under the standard single-edit protocol, where each run rewrites one factual association $\langle s,r,o\rangle \!\rightarrow\! \langle s,r,o^*\rangle$ and quantifies both the success of the rewrite and the unintended spillover to other memories.

Our primary benchmark is \textbf{COUNTERFACT\_RS}~\cite{Liu_Liu_Gu_Lin_Ma_Xiang_Wang_2025}, a relation-sensitive evaluation suite derived from \textsc{CounterFact}, on which we sample 2,000 edits and repeat evaluation five times, reporting mean and standard deviation. \textsc{COUNTERFACT\_RS} provides a controlled decomposition of editing quality into \emph{Efficacy} (whether the rewritten object is preferred under rewriting prompts), \emph{Subject Specificity} (\textsc{S-Spec.}, whether the new object spuriously transfers to neighborhood subjects $\langle s',r\rangle$), and \emph{Relation Specificity} (\textsc{R-Spec.}, whether the new object leaks to other relations of the same subject $\langle s,r'\rangle$). Because our goal is to eliminate Subject-Dominant Memory Interference, we treat \textsc{R-Spec.} as the key controllability metric, while also reporting a Fluency score that captures the quality of generated continuations under the benchmark protocol. 

To complement the evaluation with a broader subject-dominant memory interference assessment, we also evaluate on the \textbf{EVOKE}~\cite{zhang2025uncoveringoverfittinglargelanguage} benchmark and report corresponding metrics: \textsc{DP} (Deterioration Probability) to measure how often unrelated answers are disrupted after editing, \textsc{CAP} (Correct Answer Probability) to track overall correctness, and \textsc{EOS} (Editing Overfit Score) to capture over-confident collapse, serving as a primary indicator of the model’s specificity and overall performance; lower \textsc{DP} and higher \textsc{EOS} indicate stronger resistance to interference.

\paragraph{Baselines and Model Architectures.}
We compare against representative locate-then-edit editors, including \textbf{ROME}~\cite{meng2022locating}, \textbf{MEMIT}~\cite{Meng2022MassEditingMI}, and \textbf{AlphaEdit}~\cite{fang2024alphaeditnullspaceconstrainedknowledge}, as well as recent mitigation-oriented variants such as \textbf{RETS}~\cite{Liu_Liu_Gu_Lin_Ma_Xiang_Wang_2025} and \textbf{ROME-LTI}~\cite{zhang2025uncoveringoverfittinglargelanguage}. Experiments span multiple LLM families and scales, including GPT2-XL (1.5B), GPT-J (6B)~\cite{gpt-j}, and Qwen2.5 (7B/14B). In particular, we report the main results for \textbf{ROME-CPA} because \textsc{ROME} is the representative and most widely adopted single-edit backbone, and it cleanly decomposes editing into (i) optimizing the target value representation and (ii) a closed-form rank-one weight update at the located MLP module. This decomposition makes the shortcut pathology and our trajectory-alignment intervention directly comparable. We nevertheless verify portability by transplanting \textsc{CPA} to other locate-then-edit methods in subsequent experiments.

\begin{table*}[ht]
\centering
\small
\renewcommand{\arraystretch}{1.1}
\setlength{\tabcolsep}{2.2mm}
\begin{tabular}{l cccc ccc}
\hline \toprule
\multirow{2}{*}{\textbf{Method}} & \multicolumn{4}{c}{\textbf{COUNTERFACT\_RS}} & \multicolumn{3}{c}{\textbf{EVOKE}} \\
\cmidrule(lr){2-5} \cmidrule(lr){6-8}
& Eff.$\uparrow$ & S-Spec.$\uparrow$ & R-Spec.$\uparrow$ & Flue.$\uparrow$ & DP$\downarrow$ & CAP$\uparrow$ & EOS$\uparrow$ \\
\midrule
\multicolumn{8}{l}{\textit{\textbf{GPT2-XL (1.5B)}}} \\
ROME & 99.8{\color{gray}\scriptsize($\pm$0.0)} & 75.4{\color{gray}\scriptsize($\pm$0.0)} & \textcolor{red}{45.7}{\color{gray}\scriptsize($\pm$0.2)} & 623.4{\color{gray}\scriptsize($\pm$0.0)} & 39.1{\color{gray}\scriptsize($\pm$0.3)} & 2.8{\color{gray}\scriptsize($\pm$0.1)} & 21.6{\color{gray}\scriptsize($\pm$0.2)} \\
MEMIT & 94.0{\color{gray}\scriptsize($\pm$0.1)} & \underline{76.4}{\color{gray}\scriptsize($\pm$0.0)} & \textcolor{red}{61.5}{\color{gray}\scriptsize($\pm$0.5)} & \textbf{627.4}{\color{gray}\scriptsize($\pm$0.2)} & 38.0{\color{gray}\scriptsize($\pm$1.0)} & 1.9{\color{gray}\scriptsize($\pm$0.1)} & 10.9{\color{gray}\scriptsize($\pm$0.3)} \\
AlphaEdit & 99.8{\color{gray}\scriptsize($\pm$0.0)} & 76.2{\color{gray}\scriptsize($\pm$0.0)} & \textcolor{red}{41.2}{\color{gray}\scriptsize($\pm$0.3)} & \underline{623.4}{\color{gray}\scriptsize($\pm$0.1)} & 34.5{\color{gray}\scriptsize($\pm$0.2)} & 2.1{\color{gray}\scriptsize($\pm$0.0)} & 15.5{\color{gray}\scriptsize($\pm$0.2)} \\
RETS & \textbf{100.0}{\color{gray}\scriptsize($\pm$0.0)} & 67.9{\color{gray}\scriptsize($\pm$0.4)} & \textbf{78.7}{\color{gray}\scriptsize($\pm$0.5)} & \textcolor{red}{585.1}{\color{gray}\scriptsize($\pm$5.9)} & \textbf{5.2}{\color{gray}\scriptsize($\pm$0.8)} & \textbf{4.8}{\color{gray}\scriptsize($\pm$0.0)} & \textbf{61.4}{\color{gray}\scriptsize($\pm$2.0)} \\
ROME-LTI & \underline{99.9}{\color{gray}\scriptsize($\pm$0.0)} & 75.7{\color{gray}\scriptsize($\pm$0.1)} & 57.2{\color{gray}\scriptsize($\pm$0.3)} & 622.0{\color{gray}\scriptsize($\pm$0.2)} & 17.7{\color{gray}\scriptsize($\pm$0.2)} & \underline{3.6}{\color{gray}\scriptsize($\pm$0.1)} & 31.4{\color{gray}\scriptsize($\pm$0.4)} \\
\rowcolor{gray!10} \textbf{ROME-CPA (Ours)} & 98.7{\color{gray}\scriptsize($\pm$0.1)} & \textbf{76.5}{\color{gray}\scriptsize($\pm$0.0)} & \underline{72.2}{\color{gray}\scriptsize($\pm$0.7)} & 619.7{\color{gray}\scriptsize($\pm$0.3)} & \underline{5.7}{\color{gray}\scriptsize($\pm$0.1)} & 2.5{\color{gray}\scriptsize($\pm$0.1)} & \underline{36.5}{\color{gray}\scriptsize($\pm$0.1)} \\
\midrule
\multicolumn{8}{l}{\textit{\textbf{GPT-J (6B)}}} \\
ROME & \textbf{100.0}{\color{gray}\scriptsize($\pm$0.0)} & 79.3{\color{gray}\scriptsize($\pm$0.0)} & \textcolor{red}{52.8}{\color{gray}\scriptsize($\pm$0.4)} & 621.0{\color{gray}\scriptsize($\pm$0.2)} & 53.1{\color{gray}\scriptsize($\pm$0.4)} & 1.9{\color{gray}\scriptsize($\pm$0.0)} & 6.1{\color{gray}\scriptsize($\pm$0.1)} \\
MEMIT & \textbf{100.0}{\color{gray}\scriptsize($\pm$0.0)} & 81.1{\color{gray}\scriptsize($\pm$0.0)} & \textcolor{red}{66.8}{\color{gray}\scriptsize($\pm$0.3)} & \textbf{621.5}{\color{gray}\scriptsize($\pm$0.1)} & 38.8{\color{gray}\scriptsize($\pm$0.3)} & 2.2{\color{gray}\scriptsize($\pm$0.1)} & 12.3{\color{gray}\scriptsize($\pm$0.2)} \\
AlphaEdit & \underline{99.8}{\color{gray}\scriptsize($\pm$0.1)} & \underline{81.9}{\color{gray}\scriptsize($\pm$0.0)} & \textcolor{red}{60.4}{\color{gray}\scriptsize($\pm$0.2)} & \underline{621.4}{\color{gray}\scriptsize($\pm$0.1)} & 38.2{\color{gray}\scriptsize($\pm$0.2)} & 2.5{\color{gray}\scriptsize($\pm$0.0)} & 14.8{\color{gray}\scriptsize($\pm$0.1)} \\
RETS & \textbf{100.0}{\color{gray}\scriptsize($\pm$0.0)} & \textcolor{red}{61.4}{\color{gray}\scriptsize($\pm$1.9)} & 81.5{\color{gray}\scriptsize($\pm$0.7)} & \textcolor{red}{530.6}{\color{gray}\scriptsize($\pm$17.1)} & 30.1{\color{gray}\scriptsize($\pm$0.6)} & \textbf{4.2}{\color{gray}\scriptsize($\pm$0.2)} & 30.7{\color{gray}\scriptsize($\pm$0.5)} \\
ROME-LTI & \underline{99.8}{\color{gray}\scriptsize($\pm$0.0)} & 80.7{\color{gray}\scriptsize($\pm$0.1)} & \underline{82.5}{\color{gray}\scriptsize($\pm$0.3)} & 618.8{\color{gray}\scriptsize($\pm$0.1)} & \underline{9.0}{\color{gray}\scriptsize($\pm$0.5)} & \underline{3.5}{\color{gray}\scriptsize($\pm$0.1)} & \underline{32.5}{\color{gray}\scriptsize($\pm$0.5)} \\
\rowcolor{gray!10} \textbf{ROME-CPA (Ours)} & 99.6{\color{gray}\scriptsize($\pm$0.0)} & \textbf{82.0}{\color{gray}\scriptsize($\pm$0.0)} & \textbf{83.5}{\color{gray}\scriptsize($\pm$0.3)} & 618.6{\color{gray}\scriptsize($\pm$0.2)} & \textbf{7.8}{\color{gray}\scriptsize($\pm$0.7)} & 3.1{\color{gray}\scriptsize($\pm$0.1)} & \textbf{37.3}{\color{gray}\scriptsize($\pm$1.3)} \\
\midrule
\multicolumn{8}{l}{\textit{\textbf{Qwen2.5 (7B)}}} \\
ROME & 99.8{\color{gray}\scriptsize($\pm$0.0)} & 82.2{\color{gray}\scriptsize($\pm$0.0)} & 70.7{\color{gray}\scriptsize($\pm$0.2)} & \underline{625.5}{\color{gray}\scriptsize($\pm$0.1)} & 16.6{\color{gray}\scriptsize($\pm$0.2)} & 9.3{\color{gray}\scriptsize($\pm$0.1)} & 42.6{\color{gray}\scriptsize($\pm$0.3)} \\
MEMIT & \textbf{100.0}{\color{gray}\scriptsize($\pm$0.0)} & \textcolor{red}{67.3}{\color{gray}\scriptsize($\pm$0.2)} & \textcolor{red}{64.7}{\color{gray}\scriptsize($\pm$0.5)} & 624.4{\color{gray}\scriptsize($\pm$0.1)} & 12.2{\color{gray}\scriptsize($\pm$0.3)} & \underline{11.3}{\color{gray}\scriptsize($\pm$0.1)} & 52.6{\color{gray}\scriptsize($\pm$0.4)} \\
AlphaEdit & \textbf{100.0}{\color{gray}\scriptsize($\pm$0.0)} & 83.6{\color{gray}\scriptsize($\pm$0.0)} & \textcolor{red}{65.0}{\color{gray}\scriptsize($\pm$0.7)} & \textbf{626.3}{\color{gray}\scriptsize($\pm$0.1)} & 20.8{\color{gray}\scriptsize($\pm$0.2)} & 8.0{\color{gray}\scriptsize($\pm$0.1)} & 33.6{\color{gray}\scriptsize($\pm$0.2)} \\
RETS & \underline{99.5}{\color{gray}\scriptsize($\pm$0.1)} & \textcolor{red}{42.7}{\color{gray}\scriptsize($\pm$0.4)} & \textcolor{red}{52.9}{\color{gray}\scriptsize($\pm$0.3)} & \textcolor{red}{540.3}{\color{gray}\scriptsize($\pm$7.4)} & 10.2{\color{gray}\scriptsize($\pm$0.1)} & \textbf{14.1}{\color{gray}\scriptsize($\pm$0.2)} & \underline{70.5}{\color{gray}\scriptsize($\pm$0.6)} \\
ROME-LTI & \underline{99.5}{\color{gray}\scriptsize($\pm$0.1)} & \underline{83.7}{\color{gray}\scriptsize($\pm$0.3)} & \underline{84.6}{\color{gray}\scriptsize($\pm$0.1)} & 624.5{\color{gray}\scriptsize($\pm$0.3)} & \underline{4.5}{\color{gray}\scriptsize($\pm$0.1)} & 6.6{\color{gray}\scriptsize($\pm$0.1)} & 52.8{\color{gray}\scriptsize($\pm$0.5)} \\
\rowcolor{gray!10} \textbf{ROME-CPA (Ours)} & 96.2{\color{gray}\scriptsize($\pm$0.1)} & \textbf{84.0}{\color{gray}\scriptsize($\pm$0.5)} & \textbf{92.2}{\color{gray}\scriptsize($\pm$0.4)} & 623.9{\color{gray}\scriptsize($\pm$0.2)} & \textbf{0.8}{\color{gray}\scriptsize($\pm$0.0)} & 8.5{\color{gray}\scriptsize($\pm$0.1)} & \textbf{76.2}{\color{gray}\scriptsize($\pm$0.1)} \\
\midrule
\multicolumn{8}{l}{\textit{\textbf{Qwen2.5 (14B)}}} \\
ROME & \textbf{100.0}{\color{gray}\scriptsize($\pm$0.0)} & \textcolor{red}{61.3}{\color{gray}\scriptsize($\pm$0.1)} & \textcolor{red}{56.9}{\color{gray}\scriptsize($\pm$0.3)} & \underline{624.0}{\color{gray}\scriptsize($\pm$0.1)} & 18.1{\color{gray}\scriptsize($\pm$0.4)} & 9.4{\color{gray}\scriptsize($\pm$1.1)} &  45.3{\color{gray}\scriptsize($\pm$0.6)}\\
MEMIT & \textbf{100.0}{\color{gray}\scriptsize($\pm$0.0)} & \underline{79.8}{\color{gray}\scriptsize($\pm$0.0)} & \textcolor{red}{66.7}{\color{gray}\scriptsize($\pm$0.2)} & \textbf{625.4}{\color{gray}\scriptsize($\pm$0.1)} & 16.9{\color{gray}\scriptsize($\pm$0.2)} & \textbf{15.3}{\color{gray}\scriptsize($\pm$0.1)} & 48.9{\color{gray}\scriptsize($\pm$0.3)} \\
RETS & 92.0{\color{gray}\scriptsize($\pm$0.0)} & 73.6{\color{gray}\scriptsize($\pm$1.1)} & \underline{83.2}{\color{gray}\scriptsize($\pm$0.5)} & \textcolor{red}{591.5}{\color{gray}\scriptsize($\pm$6.3)} & 33.8{\color{gray}\scriptsize($\pm$0.4)} & 7.9{\color{gray}\scriptsize($\pm$0.1)} & 29.0{\color{gray}\scriptsize($\pm$0.2)} \\
ROME-LTI & \textbf{100.0}{\color{gray}\scriptsize($\pm$0.0)} & 70.5{\color{gray}\scriptsize($\pm$0.2)} & 78.1{\color{gray}\scriptsize($\pm$0.1)} & 623.4{\color{gray}\scriptsize($\pm$0.3)} & \underline{16.4}{\color{gray}\scriptsize($\pm$0.3)} & 11.6{\color{gray}\scriptsize($\pm$0.1)} & \underline{59.8}{\color{gray}\scriptsize($\pm$0.4)} \\
\rowcolor{gray!10} \textbf{ROME-CPA (Ours)} & \underline{99.5}{\color{gray}\scriptsize($\pm$0.1)} & \textbf{83.4}{\color{gray}\scriptsize($\pm$0.5)} & \textbf{84.5}{\color{gray}\scriptsize($\pm$0.4)} & 620.5{\color{gray}\scriptsize($\pm$0.2)} & \textbf{6.9}{\color{gray}\scriptsize($\pm$0.2)} & \underline{12.7}{\color{gray}\scriptsize($\pm$0.6)} & \textbf{62.8}{\color{gray}\scriptsize($\pm$0.3)} \\
\bottomrule \hline
\end{tabular}

\caption{Performance comparison on COUNTERFACT\_RS and EVOKE datasets. We explicitly focus on the Subject-Dominant Memory Interference of locate-then-edit knowledge editing, which is mainly measured by R-Specificity (R-Spec.) on COUNTERFACT\_RS, and by DP, CAP, and EOS on EVOKE. \textcolor{red}{Red values} indicate severe low values for COUNTERFACT\_RS. Standard deviations are in parentheses. \textbf{Bold} indicates the best result, and \underline{underline} indicates the second-best result. Our CPA method effectively mitigates interference while achieving least side-effects across benchmarks. All metrics except Fluency are reported in percentage (\%).}
\label{tab:main}
\end{table*}

\paragraph{Implementation Details.}
For each backbone, we follow the commonly adopted layer choices for locate-then-edit updates and select a later bottleneck layer $l_a>l$ for relation anchoring. Concretely, for 48-layer models we edit the 18th and 19th MLP layers and anchor at $l_a\!\in\!\{19,20\}$ (GPT2-XL) or $l_a\!=\!28$ (Qwen2.5 14B), while for 28-layer models we edit layer 6 and anchor at $l_a\!=\!11$ (GPT-J and Qwen2.5 7B). More details can be referred to the Appendix.

\subsection{Main Results}
\label{sec:main_results}

Table~\ref{tab:main} summarizes the results on COUNTERFACT\_RS and EVOKE across multiple LLM backbones.

\paragraph{CounterFact\_RS.}
Across all evaluated models, CPA consistently improves relation specificity (R-Spec.), indicating a substantial reduction of Subject-Dominant Memory Interference. Concretely, compared with ROME, CPA raises R-Spec. from 45.7\% to 72.2\% on GPT2-XL, from 52.8\% to 83.5\% on GPT-J, from 70.7\% to 92.2\% on Qwen2.5 7B, and from 56.9\% to 84.5\% on Qwen2.5 14B, while maintaining high editing success (Eff.) and strong subject specificity (S-Spec.). Notably, this gain does not come from overly restrictive constraints: fluency remains comparable to ROME/MEMIT, whereas methods with aggressive constraints (e.g., RETS) can exhibit pronounced fluency degradation and less stable cross-model behavior.

\paragraph{EVOKE.}
EVOKE directly probes collateral corruption and shortcut-induced overfitting. CPA achieves the lowest (or second-lowest) damage proxy DP and the best EOS on most settings, suggesting that the edited behavior is less dominated by a subject-only shortcut and induces fewer unintended changes. While RETS can reach strong CAP/EOS on some settings, it often does so with a notable trade-off in fluency and/or robustness; in contrast, CPA provides a more balanced improvement profile across architectures.

Overall, the main results demonstrate that CPA provides a reliable remedy for Subject-Dominant Memory Interference across both benchmarks and model families. CPA consistently yields large gains in relation specificity and further reduces detailed damage indicators on EVOKE, while preserving (and often saturating) editing efficacy and maintaining competitive fluency. These findings support our central claim: anchoring the optimization to a relation-aware causal bottleneck and aligning the update trajectory to reconstruct this anchor prevents causal path collapse, enabling edits that are both effective and controllable.

\subsection{Mechanistic Validation}
\label{sec:mech_validation}

\begin{figure}[t]
    \centering
    \includegraphics[width=0.47\linewidth]{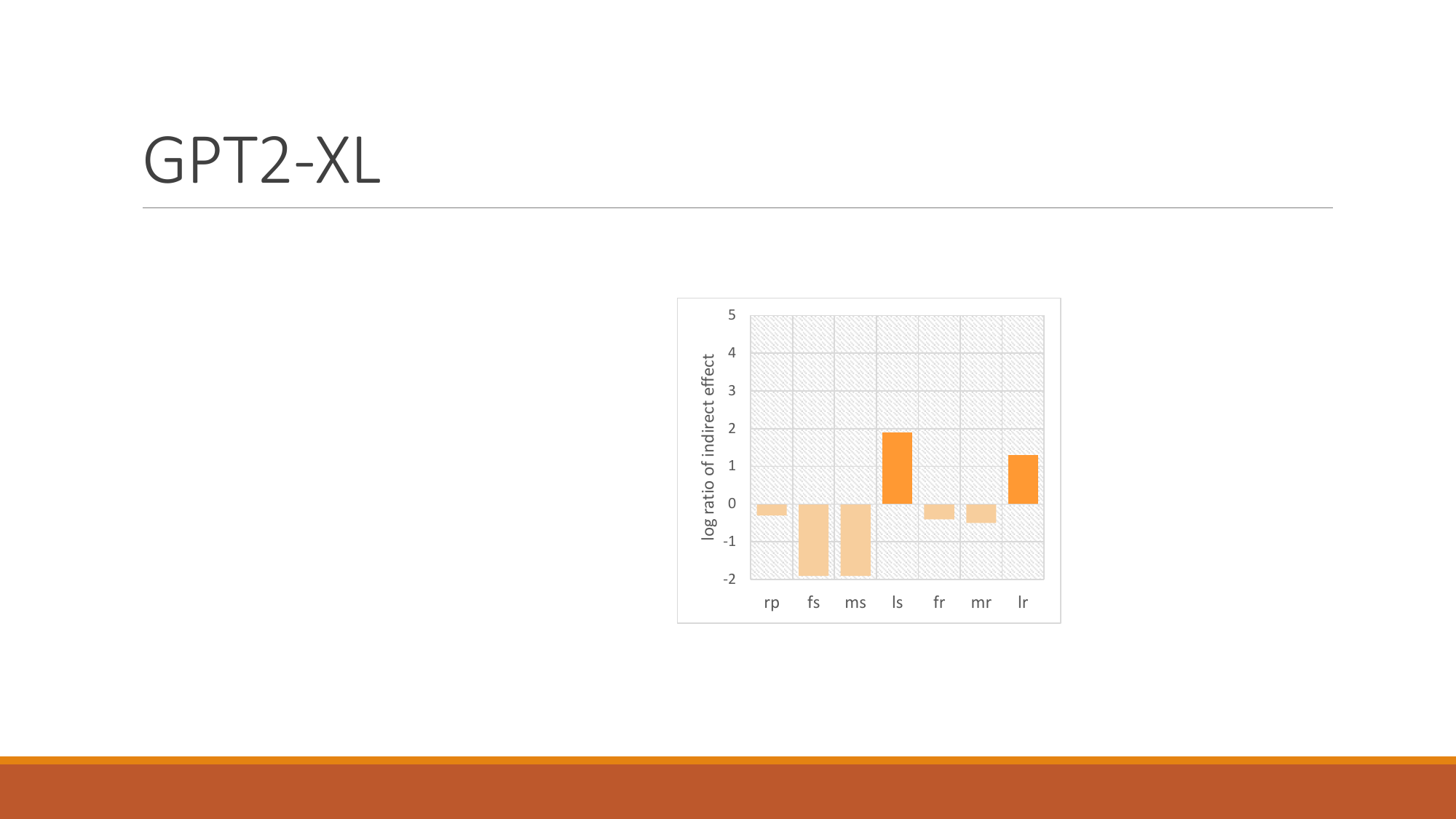}
    \includegraphics[width=0.47\linewidth]{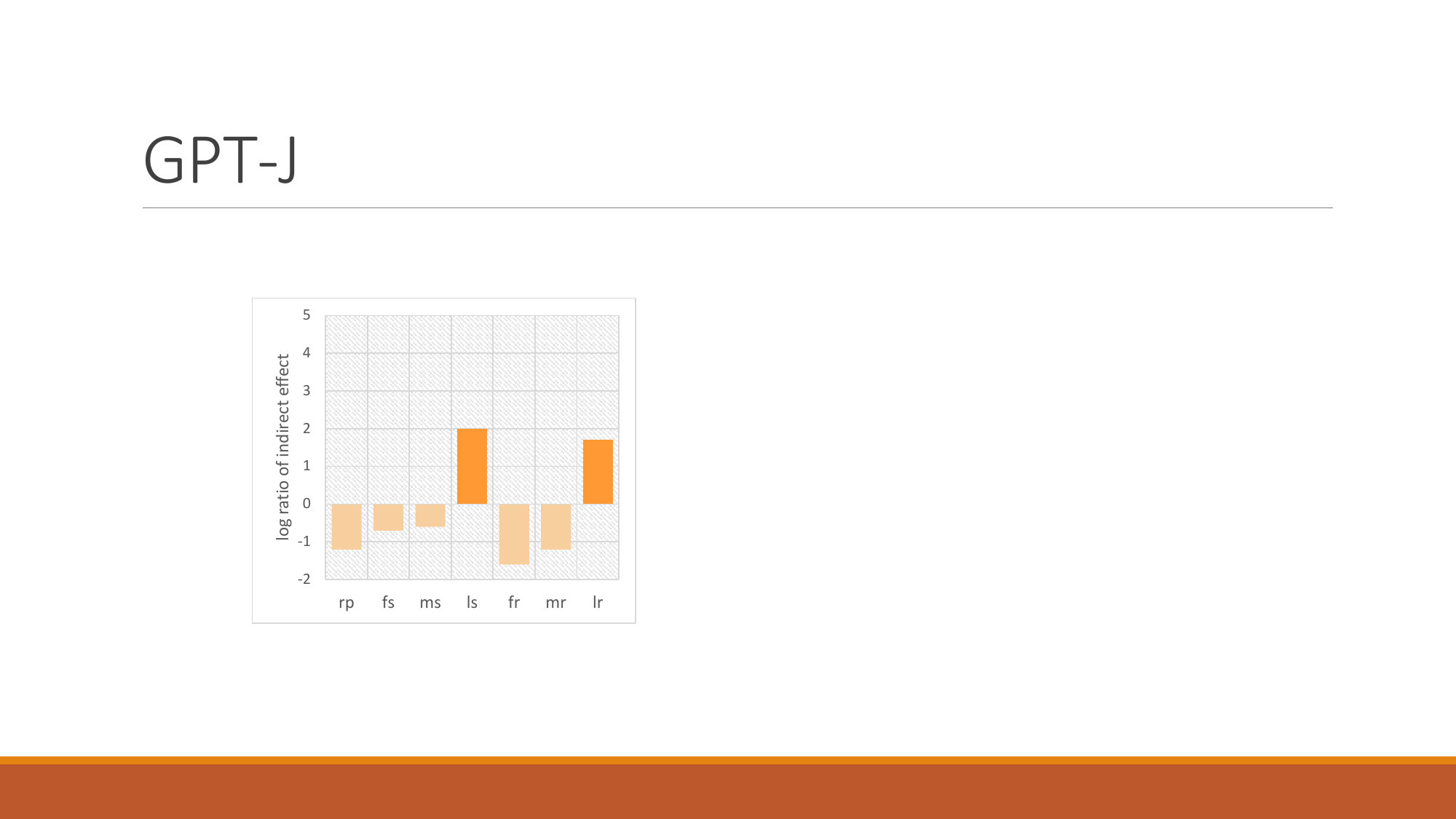}
    \caption{Mechanistic validation of eliminating the Subject Dominant Shortcut. Left: GPT2-XL (1.5B). Right: Qwen2.5 (7B).}
    \label{fig_max_rie_top}
\end{figure}

Beyond aggregate metrics, we verify that CPA indeed restores the relational causal pathway rather than merely suppressing outputs. Following Section~\ref{sec:shortcut_issue}, we re-run causal tracing and compute the maximum Ratio of Indirect Effect (RIE) over layers for each token position (Fig.~\ref{fig_max_rie_top}). The key signature of shortcut elimination is that the relative causal gain at the last relation position becomes closer to that at the last subject position, indicating that the edited prediction is mediated by relation-aware features instead of a subject-only bypass. This mechanistic evidence aligns with the consistent improvements in relation specificity and the reductions in EVOKE side-effect indicators.

\begin{figure}[t]
    \centering
    \includegraphics[width=0.48\textwidth]{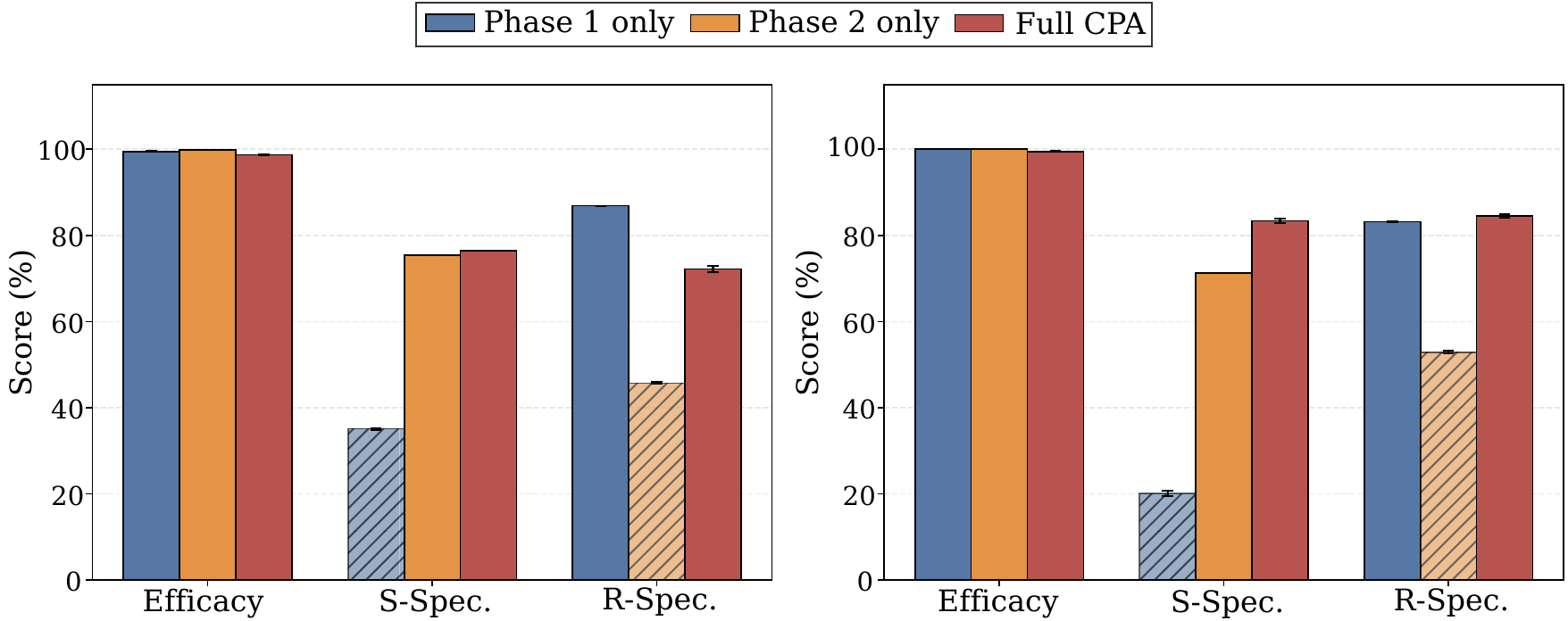}
    \caption{Ablation study of ROME-CPA on COUNTERFACT\_RS. Left: GPT2-XL (1.5B). Right: Qwen2.5 (14B).}
    \label{fig:ablation}
\end{figure}

\subsection{Ablation Study}
\label{sec:ablation}

We conduct ablation experiments on 2,000 edits from COUNTERFACT\_RS to disentangle the contribution of each phase in \textbf{Causal Path Alignment (CPA)}. We compare three configurations under identical prompts, stopping criteria, and hyperparameters: \textbf{(1) Phase 1 only} performs \emph{relation anchoring} by optimizing the bottleneck activation $h_r$ at the last-relation token position to minimize $-\log P(o^* \mid h_r)$, but omits the subsequent trajectory alignment; \textbf{(2) Phase 2 only} reduces to the conventional single-phase target construction by directly optimizing the injected subject value vector $v_s$ at the last-subject token position with the standard objective $-\log P(o^* \mid v_s)$; and \textbf{(3) Full CPA} executes the complete two-phase procedure (relation anchoring followed by trajectory alignment).

Results in Fig.~\ref{fig:ablation} confirm the synergistic necessity of both phases. We observe a distinct trade-off: \textbf{Phase~1-only} secures relation specificity but suffers from catastrophic subject over-generalization (S-Spec. drops to $\sim$20--35\%), whereas \textbf{Phase~2-only} restores subject focus but reverts to the shortcut pathology, degrading relation specificity to $\sim$46\%. Only the full CPA reconciles these conflicting objectives, achieving an optimal balance (e.g., Qwen2.5: 83.4\% S-Spec., 84.5\% R-Spec.) while maintaining near-saturated efficacy.

\subsection{Plug-and-Play Generalization to Other Editors}
\label{sec:plugin}

CPA is editor-agnostic: it constrains \emph{how} the target representation is optimized, and thus can be transplanted to other locate-then-edit pipelines without altering their localization or weight-update solvers. To test this, we integrate CPA into MEMIT and AlphaEdit by replacing their original target-vector optimization with CPA while keeping their original layer selections and update rules. As shown in Table~\ref{tab:plugin_generalization}, CPA substantially improves R-Spec. for both editors on GPT2-XL and Qwen2.5 7B (e.g., MEMIT: 61.5\%$\rightarrow$79.7\% and 64.7\%$\rightarrow$90.6\%; AlphaEdit: 41.2\%$\rightarrow$68.4\% and 65.0\%$\rightarrow$88.9\%), while largely preserving efficacy and S-Spec. This indicates that the shortcut pathology stems from the unconstrained target-vector optimization, and CPA provides a general remedy by anchoring the optimization trajectory to relation-aware causal states.

\begin{table}[t]
\centering
\renewcommand{\arraystretch}{1.1} 
\setlength{\tabcolsep}{2.0mm}

\begin{tabular}{l l c c c}
\toprule
\textbf{Model} & \textbf{Method} & \textbf{Eff.}$\uparrow$ & \textbf{S-Spec.}$\uparrow$ & \textbf{R-Spec.}$\uparrow$ \\
\midrule
\multirow{4}{*}{\shortstack[l]{\textit{GPT2-XL}\\(1.5B)}} 
 & MEMIT & 94.0 & 76.4 & 61.5 \\
 & \cellcolor{gray!15}+ \textbf{CPA} & \cellcolor{gray!15}92.8 & \cellcolor{gray!15}76.3 & \cellcolor{gray!15}\textbf{79.7} \small{(+18.2)} \\
 & AlphaEdit & 99.8 & 76.2 & 41.2 \\
 & \cellcolor{gray!15}+ \textbf{CPA} & \cellcolor{gray!15}97.7 & \cellcolor{gray!15}76.2 & \cellcolor{gray!15}\textbf{68.4} \small{(+27.2)} \\
\midrule
\multirow{4}{*}{\shortstack[l]{\textit{Qwen2.5}\\(7B)}} 
 & MEMIT & 100.0 & 67.3 & 64.7 \\
 & \cellcolor{gray!15}+ \textbf{CPA} & \cellcolor{gray!15}93.8 & \cellcolor{gray!15}82.7 & \cellcolor{gray!15}\textbf{90.6} \small{(+25.9)} \\
 & AlphaEdit & 100.0 & 83.6 & 65.0 \\
 & \cellcolor{gray!15}+ \textbf{CPA} & \cellcolor{gray!15}92.0 & \cellcolor{gray!15}82.5 & \cellcolor{gray!15}\textbf{88.9} \small{(+23.9)} \\
\bottomrule
\end{tabular}
\caption{Plug-and-Play generalization of CPA on MEMIT and AlphaEdit. All metrics are reported in percentage (\%).}
\label{tab:plugin_generalization}
\end{table}

\section{Conclusion}
In this work, we diagnose the \textit{Subject-Dominant Shortcut}—a pathology in locate-then-edit methods where optimization decouples predictions from relational context, causing severe subject-dominant memory interference. To rectify this, we propose Causal Path Alignment (CPA), a framework that anchors parameter updates to valid internal causal pathways. By enforcing a conditional independence constraint via a two-stage process, CPA ensures updates are mediated by relational features rather than unconditional subject mappings. Experiments across diverse LLMs demonstrate that CPA significantly boosts relational specificity without compromising efficacy. As a model-agnostic plug-in, CPA effectively resolves the plasticity-stability trade-off, paving the way for trustworthy parametric memory updates in evolving AI agents.

\section*{Ethical Statement}
Our research focuses on addressing the challenge of outdated or erroneous knowledge encoded in large language models by developing methods to detect and correct such inaccuracies. At the same time, we are acutely aware of the ethical risks this technology entails, particularly the possibility of misuse for disseminating harmful or misleading content. To mitigate these risks, we underscore the critical importance of two key principles: first, language models must be obtained from rigorously vetted and reliable sources; second, users must exercise utmost caution and critical judgment when utilizing outputs generated by these models.

In the preparation of this manuscript, AI-assisted tools (including large language models) are employed strictly for the purpose of polishing and textual refinement to enhance clarity and organization. The results have been further controlled and edited. We retain full responsibility for the core scientific questions, methodological design, theoretical innovation, experimental analysis, and all conclusions presented. AI plays no role in the creative scientific conceptualization process.

\section*{Acknowledgements}
This work was supported by the National Natural Science Foundation of China (Nos. 62472419, 62472420) and the Enterprise Project (No. E4V06811F3).

\bibliographystyle{named}
\bibliography{ijcai26}

\input{appendix}

\end{document}

%% file: appendix.tex
\section*{A. Implementation Details}
\label{sec:appendix_a}

This appendix supplements Section 5.1 of the main text by providing comprehensive details regarding the model architectures, layer selection strategies, and hyperparameter configurations used for Causal Path Alignment (CPA).

\subsection*{A.1 Layer Selection Strategy}
Consistent with the Locate-then-Edit paradigm, CPA operates on specific Multi-Layer Perceptron (MLP) modules. We select the editing layer $l$ based on established causal tracing results which identify early-to-mid layers as decisive for subject retrieval. The crucial relation anchor layer $l_a$ ($l_a > l$) is selected to capture the relational semantics downstream of the edit. The configurations for different Large Language Models (LLMs) are as follows:

\textbf{GPT2-XL (1.5B):} This model consists of 48 layers. We perform editing at layers $l \in \{18, 19\}$. The relation anchor is established at the immediate subsequent layers, specifically $l_a = 20$, to strictly constrain the local update trajectory.

\textbf{GPT-J (6B) \& Qwen2.5 (7B):} These models possess 28 layers. We target layer $l=6$ for editing. The relation anchor is set at a deeper bottleneck layer, $l_a=11$, to allow sufficient capacity for relational feature extraction.

\textbf{Qwen2.5 (14B):} For this 48-layer model, we edit layers $l \in \{18, 19\}$. However, unlike GPT2-XL, we anchor at a deeper layer, $l_a=28$, to align with the model's deeper semantic processing of relations.

\subsection*{A.2 Optimization Hyperparameters}
CPA employs a two-phase optimization process: \textit{Phase 1 (Relation Anchoring)} and \textit{Phase 2 (Trajectory Alignment)}. We utilize the Adam optimizer with a consistent learning rate of $5e^{-1}$ and a weight decay of $0.5$ across all experiments. The optimization is halted when the loss falls below specific thresholds ($\delta$) or reaches the maximum number of epochs.

Table~\ref{tab:hyperparams} details the specific hyperparameters used for each model family, which are mostly inherited from previous work (e.g., ROME).

\begin{table}[h]
    \centering
    \small
    \renewcommand{\arraystretch}{1.2}
    \setlength{\tabcolsep}{0.4mm}
    
    \begin{tabular}{l|ccc}
        \toprule
        \textbf{Hyperparameter} & \textbf{GPT2-XL} & \textbf{GPT-J / Qwen2.5 7B} & \textbf{Qwen2.5 14B} \\
        \midrule
        \multicolumn{4}{c}{\textit{General Settings}} \\
        Learning Rate & $5e^{-1}$ & $5e^{-1}$ & $5e^{-1}$ \\
        Weight Decay & 0.5 & 0.5 & 0.5 \\
        \midrule
        \multicolumn{4}{c}{\textit{Phase 1: Relation Anchoring}} \\
        Max Epochs & 25 & 20 & 20 \\
        Loss Threshold ($\delta_1$) & $2e^{-2}$ & $2e^{-2}$ & $2e^{-2}$ \\
        $\lambda_{KL}$ (Eq. 7) & 0.0625 & 0.0625 & 0.0625 \\
        \midrule
        \multicolumn{4}{c}{\textit{Phase 2: Trajectory Alignment}} \\
        Max Epochs & 25 & 20 & 20 \\
        Loss Threshold ($\delta_2$) & $5e^{-2}$ & $5e^{-2}$ & $5e^{-2}$ \\
        $\lambda_{F}$ (Eq. 8) & 0.1 & 0.06 & 0.1 \\
        \bottomrule
    \end{tabular}
    \caption{Hyperparameter configurations for CPA across different LLM backbones. $\lambda_{KL}$ and $\lambda_{F}$ denote the weights for the KL divergence term (Phase 1) and the Frobenius norm term (Phase 2), respectively.}
    \label{tab:hyperparams}
\end{table}

\subsection*{A.4 Baseline Implementation Details}
To ensure a fair and rigorous comparison, we strictly strictly adhere to the hyperparameter settings and implementation details provided in the original open-source repositories for all baseline methods (including ROME, MEMIT, AlphaEdit, RETS and ROME-LTI). All optimization parameters for baselines, including learning rates and regularization coefficients, replicate the configurations specified in their official codebases.

The code for baseline models are released under an open-source license to promote reproducibility and community use, while the datasets are distributed under specific terms to ensure compliance with privacy and ethical guidelines.

\subsection*{A.5 Computational Resources}
All experiments in this work are conducted on a single NVIDIA A800 (80GB) GPU.

\section*{B. Extended Analysis on Causal Path Alignment}
\label{sec:appendix_b}

This appendix provides a deeper investigation into the structural and dynamic properties of the Causal Path Alignment (CPA) framework. We specifically analyze the sensitivity of the model to the position of the relation anchor and examine the optimization dynamics that justify our two-phase design.

\subsection*{B.1 Sensitivity Analysis of Relation Anchor Layer}
The selection of the relation anchor layer, denoted as $l_a$, is the primary structural hyperparameter in CPA. This layer defines the intermediate state $h_r$ that serves as the ``virtual anchor'' for the optimization trajectory. To understand its impact, we conducted a sensitivity analysis by varying $l_a$ across the post-edit layers while maintaining the editing layer $l$ constant (e.g., $l=6$ for GPT-J).

\begin{figure}[h]
    \centering
    \includegraphics[width=0.9\linewidth]{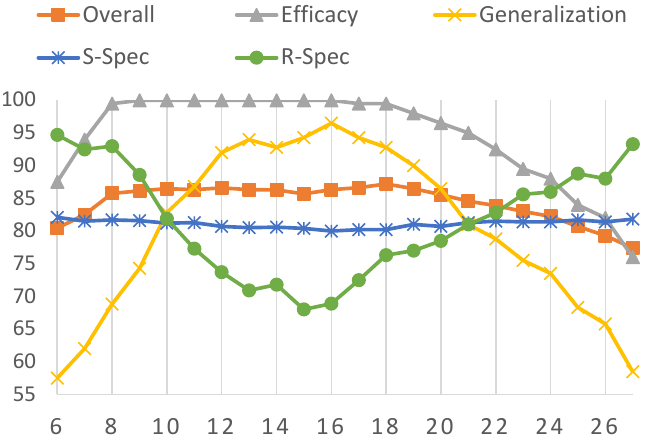}
    \caption{The performance of CPA editing with the relation anchor $h_r$ selected at different layers (x-axis) over 200 prompts for GPT-J 6B. The editing layer $l$ is fixed at layer 6.}
    \label{fig_layer_anal}
\end{figure}

As illustrated in Figure~\ref{fig_layer_anal}, the Subject Specificity (S-Spec) and Efficacy scores remain remarkably stable regardless of the anchor layer selection. This stability suggests that the relation anchor primarily constrains the relational pathway without interfering with the model's fundamental ability to retain subject-centric information or accept the new target fact. However, we observe a distinct trade-off between Relation Specificity (R-Spec) and Generalization as $l_a$ shifts depth. R-Specificity initially declines and subsequently rises as $l_a$ increases, while Generalization exhibits an inverse trend.

We attribute this phenomenon to the varying levels of semantic abstraction across the network depth. When $l_a$ is set to a deeper layer (closer to the output), the mapping from the relation anchor to the target object (optimized in Phase 1) becomes simpler, as deeper representations are already highly correlated with the output. Consequently, the optimization in Phase 1 tends to overfit the specific relation tokens in the prompt, creating a highly specific anchor that rejects unrelated relations (high R-Spec) but fails to generalize to paraphrased prompts (low Generalization). Conversely, selecting an earlier bottleneck layer for $l_a$ forces the model to learn a more robust mapping, favoring Generalization. The layer configurations detailed in Appendix A represent the optimal balance point identified between these competing objectives.

\subsection*{B.2 Optimization Dynamics and Phase Synergy}
\begin{figure}[ht]
    \centering
    \includegraphics[width=0.75\linewidth]{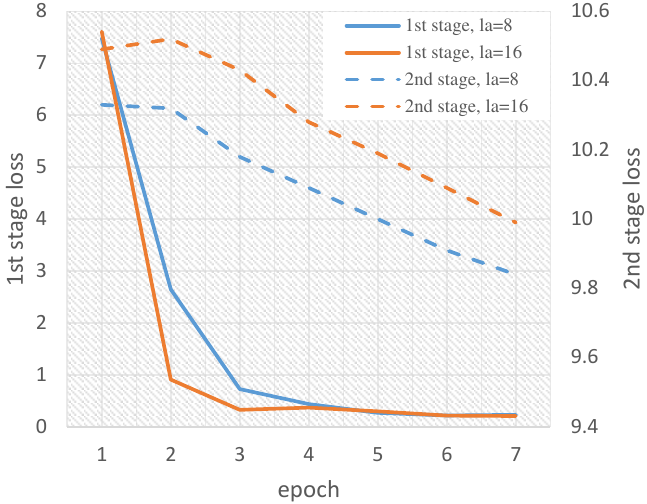}
    \caption{Average losses in the first 7 epochs of the two optimization phases for anchor layers $l_a=8$ and $l_a=16$ on GPT-J.}
    \label{fig_trade_off}
\end{figure}

To further validate the necessity of the two-phase approach, we analyzed the loss convergence behaviors during the editing process. Figure~\ref{fig_trade_off} tracks the loss curves for both \textit{Phase 1: Relation Anchoring} ($-\log P(o^*|h_r)$) and \textit{Phase 2: Trajectory Alignment} ($||\mathcal{F}[v_s] - h_r^*||_F$) over the initial epochs.

The analysis reveals a significant divergence in optimization speeds. The loss in Phase 1 decreases rapidly, particularly when the anchor layer $l_a$ is deeper (e.g., $l_a=16$), indicating that establishing a mapping from deep relation features to the target object is a relatively straightforward task. In contrast, Phase 2 exhibits a slower convergence rate, as the subject vector $v_s$ at the early edit layer must be optimized to reconstruct this downstream anchor.

This dynamic interplay validates the theoretical decomposition of our objective. If optimized jointly or without anchoring, the rapid convergence of the relation-to-object mapping would likely dominate the gradient flow, causing the model to bypass the relational constraint. By freezing the anchor state $h_r^*$ after Phase 1, CPA ensures that the complex trajectory alignment in Phase 2 has a stationary target. Furthermore, the trade-off observed in Section B.1 is mechanically visible here: as $l_a$ increases, Phase 1 converges faster (risking overfitting to relation tokens and hurting Generalization), while Phase 2 becomes slower (making it harder to strictly enforce R-Specificity). This confirms that our layer selection strategy effectively balances the difficulty between anchoring the relation and aligning the trajectory.

\section*{C. Mechanistic Visualization \& Case Studies}
\label{sec:appendix_c}

This appendix provides detailed mechanistic visualizations that corroborate the \textit{Subject-Dominant Shortcut} pathology identified in Section 3 of the main text. We present extended gradient saliency maps to demonstrate the persistence of dual retrieval pathways during optimization, and specific causal tracing case studies to illustrate the collapse of the relational pathway in standard editing methods.

\subsection*{C.1 Extended Gradient Saliency Analysis}
\label{sec:grad_saliency}
In Section 3.2 of the main text, we identify a \textit{Dual-Peak Pattern} in the gradient saliency during the first epoch of optimization, indicating that the model theoretically attempts to leverage both subject and relation features. To confirm that this is a persistent structural property rather than a transient initialization effect, we visualize the average gradient saliency maps across the entire optimization trajectory.

\begin{figure}[ht]
    \centering
    \subfigure[5 epochs]{
    \begin{minipage}[t]{0.22\textwidth}
    \centering
    \includegraphics[scale=0.25]{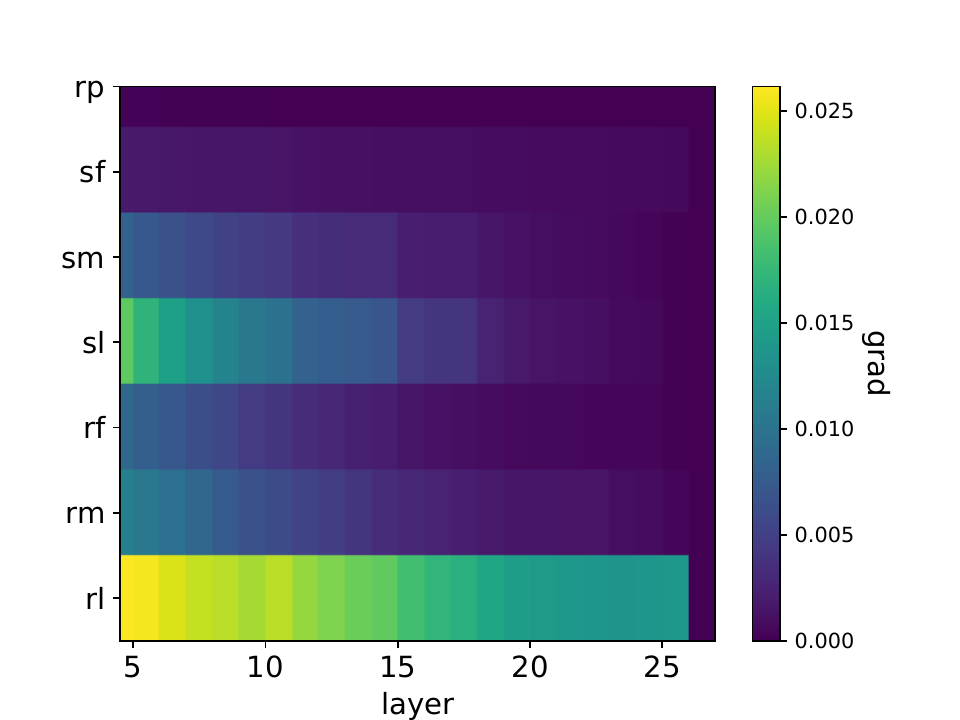}
    \end{minipage}
    }
    \subfigure[10 epochs]{
    \begin{minipage}[t]{0.22\textwidth}
    \centering
    \includegraphics[scale=0.25]{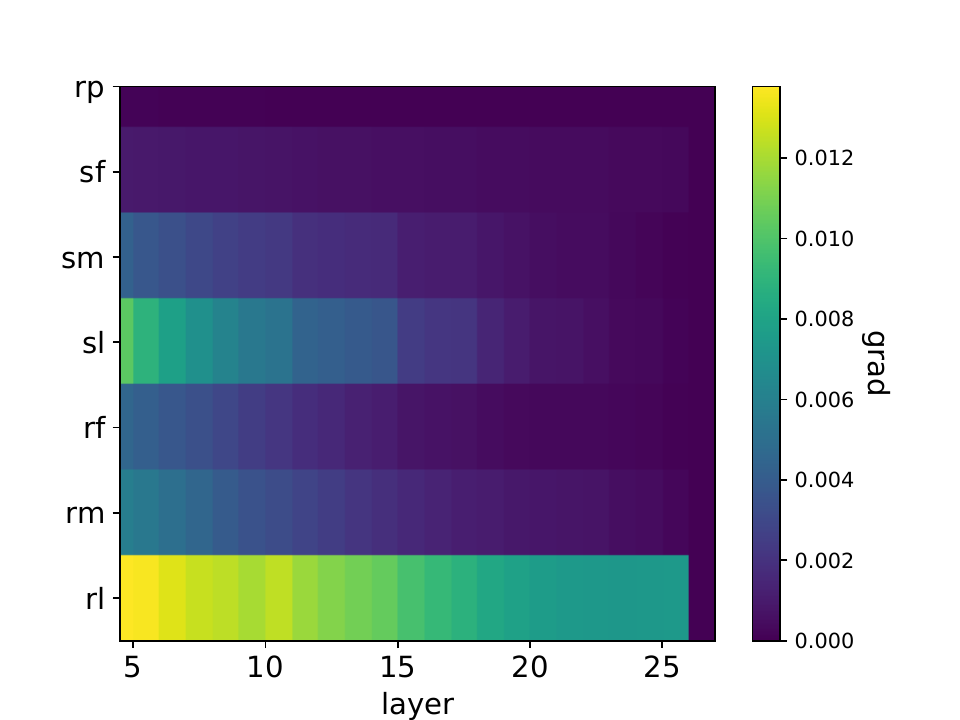}
    \end{minipage}
    }
    \subfigure[15 epochs]{
    \begin{minipage}[t]{0.22\textwidth}
    \centering
    \includegraphics[scale=0.25]{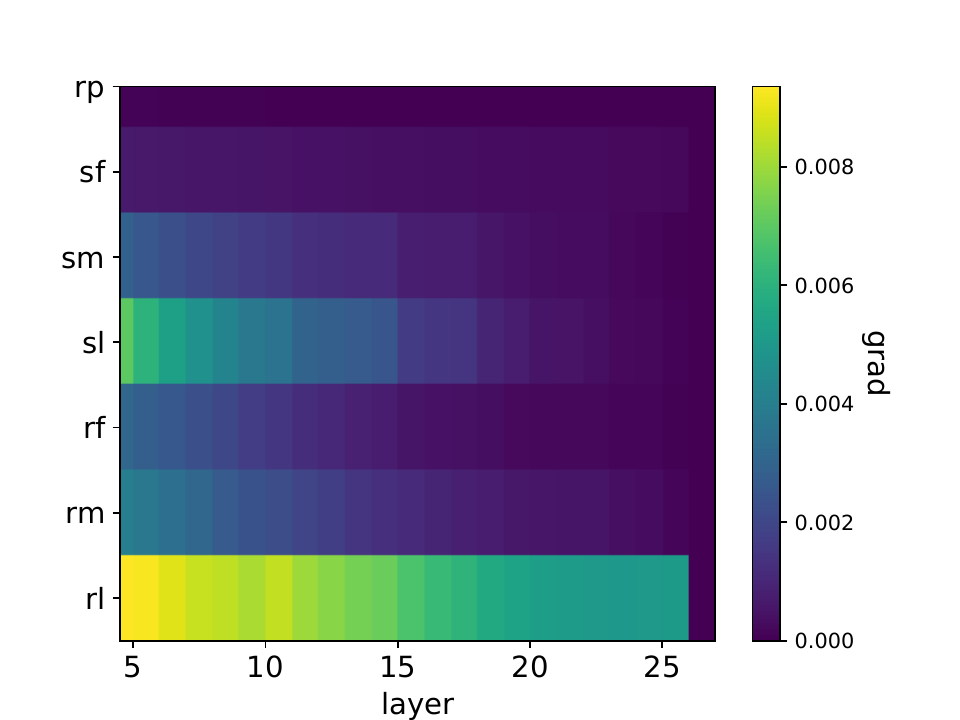}
    \end{minipage}
    }
    \subfigure[20 epochs]{
    \begin{minipage}[t]{0.22\textwidth}
    \centering
    \includegraphics[scale=0.25]{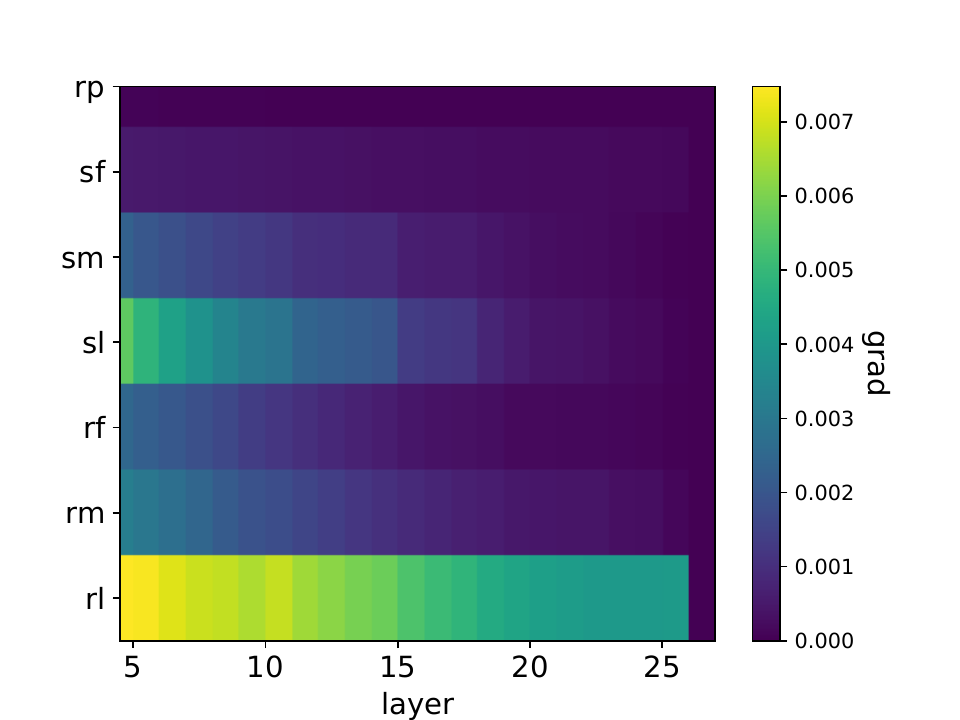}
    \end{minipage}
    }
    \caption{Average gradient saliency maps at different stages of the optimization process (5, 10, 15, and 20 epochs) for Qwen2.5 (7B). The x-axis represents layer indices, and the y-axis represents token positions. The ``Dual-Peak'' pattern at the last subject and last relation positions remains consistent throughout the entire optimization process.}
    \label{fig_grad_epochs}
\end{figure}

Figure~\ref{fig_grad_epochs} displays the gradient saliency averaged over 100 edits on Qwen2.5 (7B) at epochs 5, 10, 15, and 20 (completion).

The results show a consistent trend: the gradients remain significantly active at both the last subject token position ($p_{ls}$) and the last relation token position ($p_{lr}$) throughout the optimization process. This confirms that the relation feature ($h_r$) remains a mathematically viable pathway for minimizing the loss at every step. The failure of standard methods to utilize this path is therefore not due to a lack of gradient information, but rather due to the unconstrained objective function favoring the easier subject-shortcut, validating the necessity of our \textit{Relation Anchoring} phase.

\subsection*{C.2 Case Study of Causal Path Collapse}
\label{sec:case_study_rie}
To mechanistically visualize the ``Causal Path Collapse'' (Definition 1 in main text), we conduct a specific case study using Causal Tracing on GPT2-XL. We track the \textit{Indirect Effect (IE)} of hidden states for the query ``\textit{The mother tongue of Danielle Darrieux}''.

\begin{figure}[t]
  \centering
  \subfigure[Before Editing]{
    \begin{minipage}[t]{0.45\textwidth}
    \centering
    \includegraphics[scale=0.7]{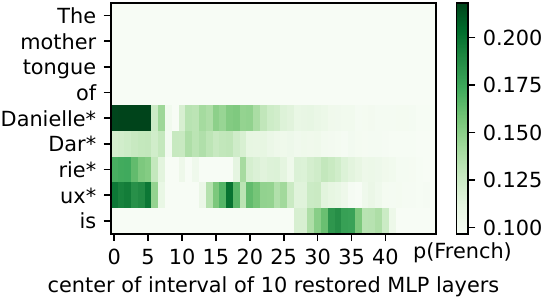}
    \end{minipage}
    }
    \subfigure[After ROME Editing]{
    \begin{minipage}[t]{0.45\textwidth}
    \centering
    \includegraphics[scale=0.7]{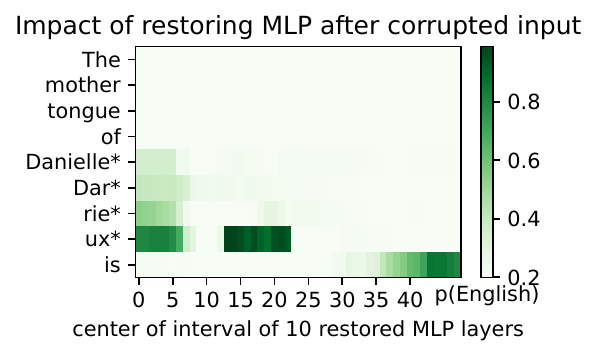}
    \end{minipage}
    }
  \caption{Causal Tracing of \textbf{Subject Tokens}: The impact of restoring MLP states after corrupting the subject ``Danielle Darrieux''. After editing (Right), the causal contribution of the subject tokens (specifically at the last subject position) explodes, increasing from $\sim$0.2 to $\sim$0.8. This indicates the model has over-learned to rely on the subject representation.}
  \label{fig_causal_case_sub}
\end{figure}

\begin{figure}[t]
  \centering
  \subfigure[Before Editing]{
    \begin{minipage}[t]{0.44\textwidth}
    \centering
    \includegraphics[scale=0.7]{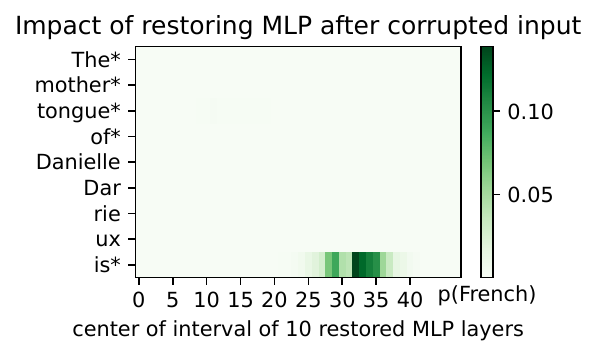}
    \end{minipage}
    }
    \subfigure[After ROME Editing]{
    \begin{minipage}[t]{0.44\textwidth}
    \centering
    \includegraphics[scale=0.7]{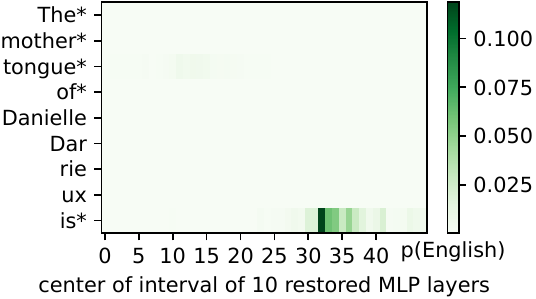}
    \end{minipage}
    }
  \caption{Causal Tracing of \textbf{Relation Tokens}: The impact of restoring MLP states after corrupting the relation ``The mother tongue of''. After editing (Right), the contribution of relation tokens fails to increase significantly and remains marginalized compared to the subject tokens. This visualizes the \textit{Causal Path Collapse}: the prediction $o^*$ becomes decoupled from the relational context.}
  \label{fig_causal_case_rel}
\end{figure}

Figure~\ref{fig_causal_case_sub} and Figure~\ref{fig_causal_case_rel} compare the causal contribution of subject and relation tokens before and after standard ROME editing.

As shown in Figure~\ref{fig_causal_case_sub}, the causal influence of the subject tokens drastically increases post-edit (max IE rises from $\approx 0.2$ to $\approx 0.8$). Conversely, Figure~\ref{fig_causal_case_rel} reveals that the relation tokens' contribution remains stagnant or even diminishes. This imbalance visually confirms that the optimization has forced the subject representation to absorb the target mapping directly ($v_s \rightarrow o^*$), bypassing the relational computation and leading to the observed memory interference.

\section*{D. Supplementary Experimental Results}
\label{sec:appendix_d}

In this section, we present supplementary experiments regarding the computational efficiency of our method and its performance on question-answering style datasets. These results provide a holistic view of the trade-offs and applicability boundaries of Causal Path Alignment (CPA).

\subsection*{D.1 Computational Overhead Analysis}
\label{sec:comp_overhead}
A natural concern with any multi-phase optimization framework is the potential for increased computational cost. To quantify this, we compare the average maximum GPU memory consumption and editing latency per sample of CPA against standard baselines (ROME, MEMIT) as well as the mitigation-oriented variant ROME-LTI. All models were evaluated in full-precision (FP32/BF16) without quantization on a single NVIDIA A800 (80GB) GPU.

\begin{figure}[ht]
    \centering
    \includegraphics[width=0.95\linewidth]{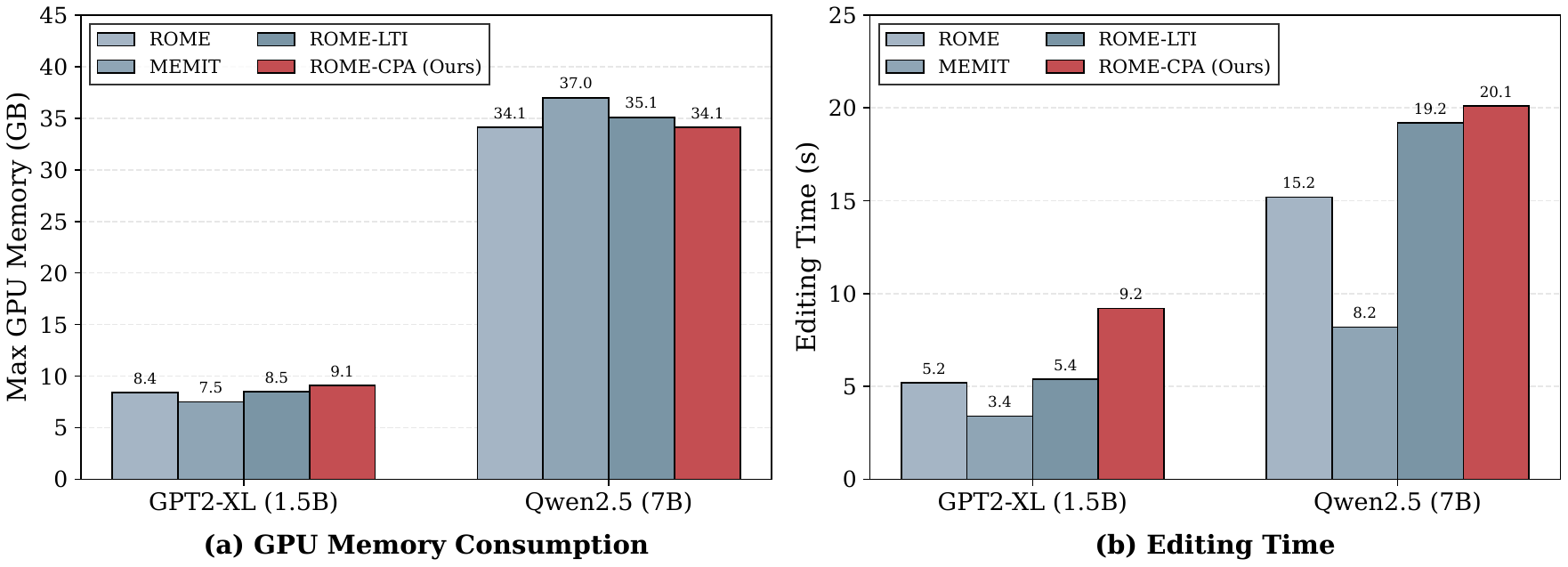}
    \caption{Comparison of average Peak GPU Memory Consumption and Average Editing Latency per sample.}
    \label{fig:comp_over}
\end{figure}

Figure~\ref{fig:comp_over} summarizes the resource consumption.

The empirical results reveal distinct characteristics regarding the time-memory trade-off. In terms of latency, CPA inevitably introduces an increase in editing time compared to single-stage methods like ROME and MEMIT (e.g., approximately 20 seconds vs. 15 seconds on Qwen2.5). This increase is a direct consequence of the sequential nature of our two-phase optimization, where the Relation Anchoring phase must complete before the Trajectory Alignment phase begins. However, when compared to ROME-LTI, another method specifically designed to mitigate overfitting through data augmentation constraints, the latency of CPA is highly comparable (20.1s vs. 19.2s on Qwen2.5). This suggests that the computational cost of ROME-CPA is within the standard acceptable range for methods that prioritize robustness and controllability over raw speed.

Regarding memory efficiency, ROME-CPA demonstrates a significant advantage. Despite the two-phase process, it introduces negligible or no additional memory overhead compared to the standard ROME baseline (e.g., identical peak memory of 34.1 GB on Qwen2.5-7B). Furthermore, it remains more memory-efficient than ROME-LTI (35.1 GB) and MEMIT (37.0 GB) in high-resource settings. This efficiency stems from the sequential execution of phases, which avoids the need to hold expanded computation graphs or augmented data batches in memory simultaneously. Therefore, while ROME-CPA incurs a modest latency cost, it maintains a minimal memory footprint, making it a practical solution for resource-constrained environments where precision is paramount.

\subsection*{D.2 Evaluation on Zero-Shot Relation Extraction (zsRE)}
\label{sec:zsre_eval}
To explore the generalization of our method across different data formats, we evaluated CPA on the Zero-Shot Relation Extraction (zsRE) dataset. Unlike COUNTERFACT which uses declarative prompts (e.g., "\emph{Lionel Messi is a citizen of}"), zsRE utilizes Question-Answering (QA) style prompts generated via back-translation (e.g., "\emph{What is the citizenship of Lionel Messi?}"). We evaluated 1,500 samples on GPT-J using standard metrics: Efficacy, Generalization, and Specificity.

Table~\ref{tab:zsre_re} presents the comparative results.

\begin{table}[t]
\centering
\renewcommand{\arraystretch}{1.2}
\setlength{\tabcolsep}{2.5mm}{
\begin{tabular}{lcccc}
\toprule
\textbf{Editor} & \textbf{Efficacy}$\uparrow$ & \textbf{Gen.}$\uparrow$ & \textbf{Specificity}$\uparrow$ \\
\midrule
Raw Model & 26.0 & 25.6 & 26.8 \\
FT-L & 68.7 & 35.0 & 26.9 \\
ROME & 99.9 & 94.2 & 26.9 \\
RETS & 93.9 & 96.2 & 26.8 \\
MEMIT & 99.8 & 87.5 & 26.9 \\
\rowcolor{gray!15} \textbf{CPA (Ours)} & 87.1 & 71.0 & 26.9 \\
\bottomrule
\end{tabular}}
\caption{Performance evaluation on the zsRE dataset (GPT-J). Note that the Specificity metric in zsRE is known to be less rigorous than in COUNTERFACT\_RS. All metrics are reported in percentage(\%).}
\label{tab:zsre_re}
\end{table}

We observe that CPA achieves lower Efficacy and Generalization scores on zsRE compared to declarative baselines like ROME. We attribute this performance gap to the structural differences in the prompts. The Locate-then-Edit paradigm (and by extension CPA) relies on the specific causal role of the \textit{last subject token} to trigger attribute retrieval. In QA-style prompts (e.g., "\emph{What was the name of...}"), the syntactic structure often places the subject in a position that may not serve as a strong causal hub for the relation feature $h_r$ in the same way declarative sentences do. Consequently, the \textit{Relation Anchoring} phase may struggle to isolate a stable relation vector $h_r$ from the question framing, leading to a less effective update.

It is also important to note that the Specificity metric in zsRE is less discerning than COUNTERFACT\_RS, resulting in indistinguishable scores across all methods (approx. 26.9). These findings suggest that while CPA is highly effective for declarative knowledge editing, future work is needed to enhance the robustness of causal path alignment for diverse prompt structures such as interrogative sentences.

\section*{E. Limitations and Future Work}
\label{sec:limitations}

While Causal Path Alignment (CPA) successfully mitigates Subject-Dominant Memory Interference, we acknowledge certain limitations that open avenues for future research.

First, as evidenced by the performance on zsRE (Appendix D.2), CPA relies on the structural assumption of the Locate-then-Edit paradigm, where the last subject token serves as a stable causal hub for attribute retrieval. This assumption holds for declarative factual statements but may weaken in interrogative forms or complex syntactic structures where the subject-relation dependency is implicit or inverted. Future work should explore \textbf{dynamic anchor identification mechanisms} that can adaptively locate relation-aware tokens in unstructured prompts.

Second, the sequential nature of our two-phase optimization inevitably incurs higher latency compared to single-stage editors. Although the memory overhead is negligible, the inference time increase (approx. $5$s per edit) may be a constraint for real-time applications requiring ultra-low latency. We plan to investigate methods to \textbf{distill the trajectory alignment constraints into a unified optimization objective}, aiming to achieve the controllability of CPA with the efficiency of single-stage methods. 

Finally, extending CPA to handle sequential batch editing in lifelong learning scenarios remains a promising direction to ensure long-term model stability.